\documentclass{ieeeaccess}
\usepackage{cite}
\usepackage{amsmath,amssymb,amsfonts}
\usepackage{graphicx}
\usepackage{textcomp}
\usepackage[T5]{fontenc}
\usepackage[utf8]{inputenc}
\usepackage[vietnamese,english]{babel}
\usepackage[normalem]{ulem}
\usepackage{float}
\usepackage{multirow}
\usepackage{caption}
\usepackage{algorithm}
\usepackage[noend]{algpseudocode}
\usepackage{makecell}
\usepackage{tikz}
\usepackage{pgfplots}
\usepackage{adjustbox}
\usepackage{placeins}

\NewSpotColorSpace{PANTONE}
  \AddSpotColor{PANTONE} {PANTONE3015C} {PANTONE\SpotSpace 3015\SpotSpace C} {1 0.3 0 0.2}
  \SetPageColorSpace{PANTONE}

\def\BibTeX{{\rm B\kern-.05em{\sc i\kern-.025em b}\kern-.08em
    T\kern-.1667em\lower.7ex\hbox{E}\kern-.125emX}}

\definecolor{color1}{HTML}{181F50}
\definecolor{imagine}{HTML}{87cfeb}
\definecolor{lightblue}{HTML}{4f81c5}
\definecolor{oxfordblue}{HTML}{1f3057}
\definecolor{seagreen}{HTML}{96bf65}
\definecolor{olivegreen}{HTML}{4b5729}
\definecolor{lightbeige}{HTML}{f5e7a1}
\definecolor{goldenyellow}{HTML}{fcc808}
\definecolor{champagne}{HTML}{e6c18d}
\definecolor{salmonpink}{HTML}{f3b28b}
\definecolor{ruby}{HTML}{982d57}
\definecolor{lightorange}{HTML}{ffc299}
\definecolor{lightgray}{HTML}{bec0b8}
\definecolor{lightviolet}{HTML}{c9a8ce}
\definecolor{shellpink}{HTML}{fbded6}
\definecolor{mediumblue}{HTML}{0000CD}
\definecolor{lilac}{HTML}{c9a8ce}
\definecolor{saxo}{HTML}{1C5B92}
\definecolor{blue2}{HTML}{99c2ff}
\definecolor{lightgray2}{HTML}{d1d1e0}
\definecolor{redpink}{HTML}{ff9999}
\definecolor{lightyellow}{HTML}{ffffb3}
\definecolor{lightgreen}{HTML}{b3ffb3}
\definecolor{customcolor}{HTML}{00cc00}

\begin{document}
\doi{10.1109/ACCESS.2020.3035701}

\title{Enhancing Lexical-Based Approach with External Knowledge for Vietnamese Multiple-Choice Machine Reading Comprehension}
\author{\uppercase{Kiet Van Nguyen}\authorrefmark{1,2}, 
\uppercase{Khiem Vinh Tran}\authorrefmark{1,2},
\uppercase{Son T. Luu}\authorrefmark{1,2},
\uppercase{Anh Gia-Tuan Nguyen}\authorrefmark{1,2}, and 
\uppercase{Ngan Luu-Thuy nguyen}\authorrefmark{1,2}
}
\address[1]{University of Information Technology, Ho Chi Minh City, Vietnam}

\address[2]{Vietnam National University, Ho Chi Minh City, Vietnam}


\markboth
{Kiet Van Nguyen \headeretal: Enhancing Lexical-Based Approach with External Knowledge for Vietnamese Multiple-Choice MRC}
{Kiet Van Nguyen \headeretal: Enhancing Lexical-Based Approach with External Knowledge for Vietnamese Multiple-Choice MRC}

\corresp{Corresponding author: Ngan Luu-Thuy Nguyen (e-mail: ngannlt@uit.edu,vn).}

\begin{abstract}
Although Vietnamese is the 17$^{th}$ most popular native-speaker language\footnote{https://en.wikipedia.org/wiki/List\_of\_languages\_by\_number\_of\_native\_speakers} in the world, there are not many research studies on Vietnamese machine reading comprehension (MRC), the task of understanding a text and answering questions about it. One of the reasons is because of the lack of high-quality benchmark datasets for this task. In this work, we construct a dataset which consists of 2,783 pairs of multiple-choice questions and answers based on 417 Vietnamese texts which are commonly used for teaching reading comprehension for elementary school pupils. In addition, we propose a lexical-based MRC method that utilizes semantic similarity measures and external knowledge sources to analyze questions and extract answers from the given text. We compare the performance of the proposed model with several baseline lexical-based and neural network-based models. Our proposed method achieves 61.81\% by accuracy, which is 5.51\% higher than the best baseline model. We also measure human performance on our dataset and find that there is a big gap between machine-model and human performances. This indicates that significant progress can be made on this task. The dataset is freely available on our website\footnote{https://sites.google.com/uit.edu.vn/uit-nlp/datasets-projects} for research purposes.
\end{abstract}


\titlepgskip=-15pt

\maketitle

\section{\textbf{Introduction}}
\label{intro}
A primary goal of computational linguistics or natural language processing is to make computers able to understand natural language texts, as well as human beings, do. One of the standard tests of natural language understanding ability requires computers to read documents and answer any questions related to their contents, resulting in different research problem settings of machine reading comprehension \cite{hermann2015,squadr2016,choi2018,10.1145/3365679,Reddy2019}. MRC can also be the extended task of question answering (QA). There are many studies on QA \cite{qa1,qa2,qa3,10.1145/3359988}, which are also the foundation for the development of MRC. Findings of this research field are implemented into various artificial intelligence applications such as next-generation search engines, AI agents, chatbots, and robots.



One common method for evaluating someone’s understanding of texts is by giving them a multiple-choice reading comprehension test. This type of test can measure abilities such as causal or counterfactual reasoning, inference among relations, or basic understanding of the world in a set of reading texts. In the past ten years, there have been many study works \cite{richardson2013,mostafazadeh2016,lai2017,khashabi2018,ostermann2018,liu2019r,lui2020} in this field. In addition to researching MRC in each language, one of the current trends in MRC is cross-lingual studies such as \cite {yuan2020enhancing,lewis2019mlqa}. Hence, the first important thing is the contribution of MRC datasets in each language. Besides, there have been research results in lexical-based approaches \cite{richardson2013,lai2017} and machine-learning-based approaches \cite{10.1145/3365679,lai2017,chen2016x,dhingra2016,zhu2018,luijcai2020}. Depending on the characteristics and size of datasets, we propose the appropriate methods to achieve better performances. 

English and Chinese are regarded as resource-rich languages when it comes to the accessibility of the tools needed to carry out communication. Still, many other languages are deemed resource-poor, and Vietnamese is one of them. Machine reading comprehension for the Vietnamese language is vital as for other languages because it is useful for non-Vietnamese speaking people to understand the question of others and answers extracted from a document or text. Vietnamese is the national language of Vietnam and is widely used by over 97 million people\footnote{https://www.worldometers.info/world-population/vietnam-population/}. Therefore, machine reading comprehension has become significant even for the Vietnamese language so that people can understand the questions and documents of people expressed in this language. The challenge of machine reading comprehension for Vietnamese has not yet been explored fully even after its extensive use; therefore, in this article, the primary focus is Vietnamese.

The integration of external sources has proven effective on a range of previous study works \cite{bi2019incorporating,10.1145/3314936} and recently, success on leveraging external knowledge to generate answers in the neural QA model \cite{bi2019incorporating}. WordNet and word embeddings are two useful external sources for a range of natural language applications. Multiple deep learning-based approaches \cite{lai2017,chen2016x,dhingra2016,zhu2018,xu2019enhancing,yang2020improved} 
have worked well when using word embeddings in multiple-choice machine reading comprehension. Because our dataset is limited in the number of questions, we aim to find solutions based on the lexical-based method when leveraging external sources in multiple-choice machine reading comprehension. Thus, our proposed method is shown in Section \ref{method} with our experiments and result analysis in Section \ref{experiment} and Section \ref{error}.

In this article, we have three main contributions described as follows.				
\begin{itemize}

\item We propose a benchmark dataset for evaluating Vietnamese multiple-choice reading comprehension task. Our dataset is the first dataset for Vietnamese multi-choice machine reading comprehension. The number of questions in our dataset is larger than that of MCTest \cite{richardson2013}, which is the English first dataset published to motivate many MRC studies. The dataset is available freely for the research community and is expected to contribute to the research development of Vietnamese machine reading comprehension. We also provide this dataset for the cross-lingual research with other similar datasets such as MCTest \cite{richardson2013}, RACE \cite{lai2017}, and C$^{3}$ \cite{sun2020investigating}.

\item We propose the lexical-based method utilizing semantic similarity and external knowledge sources for multiple-choice reading comprehension. As a result, this model achieves better accuracy than baseline models. Also, we compare this model with different baseline lexical-based and neural network-based models.

\item To gain an in-depth understanding of our proposed model, we analyze and compare its and other models' performances with different linguistic properties by quantitative analysis and visualizing their effects. Through empirical observations, researchers are given more insights and better understandings of the aspects of our proposed method on our dataset.

\end{itemize}
		 	 	 				
The rest of this paper is structured as follows. Section \ref{relate} reviews related datasets and methods. Section \ref{data} introduces the creation process and analysis of the ViMMRC dataset. Section \ref{method} presents our proposed method for Vietnamese multiple-choice machine reading comprehension. Section \ref{experiment} shows experiments and results on the dataset. Section \ref{error} describes the result analysis for these experimental results. Finally, Section \ref{conclusion} concludes the paper and discusses future work.
\section{\textbf{Literature review}}
\label{relate}

In this section, we aim to review recent datasets and techniques in machine reading comprehension. In particular, the typical MRC datasets and methods are described as follows.
\subsection{MRC datasets}
In the last decade, we have witnessed a fast growth of research interest in machine reading comprehension (MRC) and an explosion of datasets for MRC studies for popular languages like English \cite{hermann2015,squadr2016,hill2016,trischler2016,chen2016,joshi2017,welbl2018,lee2018,li2018,ma2018} and Chinese \cite{he2017dureader,shao2018drcd,cui2018span}.

In terms of types of answers, MRC datasets are divided into three categories, including extractive, abstractive, and multiple-choice. 

\begin{itemize}
    \item {\bf Extractive MRC} requires computers to locate the correct segment in a provided reading text that answers a specific question related to that text. Recently, there has been a significant increase in the construction of extractive MRC datasets with formal written texts such as SQuAD \cite{squadr2016}, CNN/Daily Mail \cite{hermann2015}, CBT \cite{hill2016}, NewsQA \cite{trischler2016}, TriviaQA \cite{joshi2017}, WIKIHOP \cite{welbl2018}, DRCD \cite{shao2018drcd}, and CMRC2018 \cite{cui2018span}. There are also datasets of which reading texts are spoken language, such as ODSQA \cite{lee2018} and Spoken SQuAD \cite{li2018} and conversation-based datasets \cite{chen2016,ma2018}. 

    \item In contrast to extractive MRC, {\bf abstractive MRC} requires computers to generate answers or synthetic summaries because answers to such questions in abstractive MRC are usually not spans in the reading text. Datasets for abstractive MRC include MS MARCO \cite{nguyen2016}, SearchQA \cite{dunn2017}, NarrativeQA \cite{kocisk2018}, and DuReader \cite{he2017dureader}.

    \item {\bf Multiple-choice MRC} includes both extractive and abstractive MRCs; however, the correct answer options are primarily abstractive. Most of the multiple-choice MRC datasets are created using crowdsourcing methods in major steps of dataset construction including generating questions, correct answer options and distractors. MCTest \cite{richardson2013},  ROCStories \cite{mostafazadeh2016}, MultiRC \cite{khashabi2018}, MCScript \cite{ostermann2018}, and COSMOS QA \cite{huang2019cosmos} are typical datasets of this type. The crowd workers also assign to each question the reasoning mechanism that is needed to figure out the answer. Apart from the basic reasoning mechanism - the matching type, a dramatic number of questions require complex reasoning mechanisms which are based on multiple sentences and require external knowledge. Other datasets are collected from examinations designed by educational experts QALD \cite{penas2014}, NTCIR-11 QA-Lab \cite{shibuki2014}, dataset from TOEFL exams \cite{tseng2016}, dataset from NY Regents 4th Grade Science exams \cite{clark2016}, and RACE \cite{lai2017}, which aim to evaluate learners.
\end{itemize}

Until now, there is not yet any dataset available for Vietnamese machine reading comprehension, which is one of the primary reasons that we would like to collect and build a dataset for the Vietnamese language processing community.

\subsection{MRC methods}
In this paper, we focus on two main types of MRC method, lexical-based approaches and neural network-based approaches. Therefore, we review the previous study works in these methodologies as follows.

\textbf{Lexical-based approaches}. The first method implemented into multiple-choice reading comprehension is the Sliding Window algorithm, a lexical-based approach developed by \cite{richardson2013}, as our first baseline model. This method was also used as a baseline in other studies
\cite{squadr2016,ostermann2018,lai2017}. Sliding Window finds an answer based on simple lexical information. Motivated by TF-IDF, this algorithm uses inverse word count as a weight of each lexical unit, and maximizes the bag-of-word similarity between the answer option and lexical units in the given reading text in a window size.

\textbf{Neural network-based approaches}. With the popularity of the neural network approach, end-to-end models such as Stanford AR \cite{chen2016x}, GA Reader \cite{dhingra2016}, HAF \cite{zhu2018}, and Co-Match \cite{wang2018} have produced promising results on multiple-choice MRC. Recently, pre-trained language models have also been added  \cite{radford2019,devlin2018}. These models do not rely on complex manually-devised features as in traditional machine learning approaches, but are able to outperform them. In this paper, we employ an end-to-end model called Co-match \cite{wang2018} with different pre-trained word embeddings as another baseline model.

Regarding to the Vietnamese language processing, there are quite a number of research works on other tasks such as parsing \cite{nguyen2014treebank,nguyen2016x,nguyen2018lstm}, part-of-speech \cite{datpos,bachpos}, named entity recognition \cite{10.1145/1316457.1316460,10.1145/2990191,nguyen2019error}, sentiment analysis \cite{van2018uit,nguyen2018deep,van2018transformation}, and question answering \cite{Nguyen_2009,van2016improving,le2018factoid}. However, to the extent of our knowledge, there are no research publications on multiple-choice machine reading comprehension. Therefore, we decide to build a new dataset of Vietnamese multiple-choice reading comprehension for the research community and evaluate MRC state-of-the-art models on our dataset.

\subsection{Semantic similarity measurement and word embeddings}
Recently, the semantic similarity measures between texts have been studied in many natural language processing applications. A range of researchers have used these measures to improve their study works \cite{10.1145/3182622,gupta2017continuous,zhou2019text}. These methods proposed for estimating the similarity between two documents include three different types, i.e., lexical matching, linguistic analysis, and semantic features. Lexical matching is not sufficiently strong and linguistic analysis also have limitations. In semantic feature approaches, a word is represented by a vector as semantic meaning before estimating similarity. 
These study works \cite{meng2013review,jiang2017wikipedia} utilized external knowledge sources to estimate the similarity of two texts. These approaches are only effective when external knowledge sources such as WordNet, word embeddings or other datasets are available for the tested domain or applications.

Word embeddings also play a significant role in machine reading comprehension. Rumelhart et al. (1986) \cite{rumelhart1986} proposed word embedding, a technique that maps each word to a vector space and can accurately capture a large proportion of syntactic and semantic relationships in text. Using pre-trained word embedding \cite{seo2017,hu2018}, there are two most common methods to represent words in machine reading comprehension models: word-level embedding and character-level embedding. However, these methods seem to be insufficient because it simply concatenates word-level and character-level embeddings; generated vectors stay the same in different contexts. To tackle these problems, Peters et al. (2018)\cite{peter2018} proposed deep contextualized word representations called ELMo which is pre-trained by language model first and fine-tuned according to the learning task. Devlin et al. (2018) \cite{devlin2018} introduced BERT, which utilizes bidirectional transformer to encode both left and right contexts to the representations. In this article, we take advantage of semantic similarity and word embeddings for enhancing external knowledge to improve the performance of multiple-choice reading  comprehension in Vietnamese.

\section{Dataset}
\label{data}
\subsection{Dataset creation}
The process of constructing the ViMMRC dataset includes three different phases: reading-text collection, multiple-choice question creation, and dataset validation. These phases are described in detail as follows.

\textbf{Reading-text collection:} We decide to focus on the reading comprehension levels at primary schools because they only require general knowledge, not too specific knowledge. We collect the Vietnamese reading texts suitable for the $1^{st}$ to $5^{th}$ graders from the subject named Vietnamese. In addition, we collect  reading comprehension tests from two reliable websites where all reading comprehension tests from $1^{st}$ to $5^{th}$ grades are made public for free of charge. As a result, 417 reading texts are gathered. 

\textbf{Multiple-choice question collection: } Questions, answer options, and correct answers are created by primary-school teachers. These questions are intended to test the reading comprehension ability of elementary learners. The teachers are asked to create at least five questions per text. Each question is accompanied by four answer options, of which only one is correct. For those texts with fewer numbers of questions or answer options, it is necessary to create more to meet the above conditions. Spelling errors are corrected. At the end of this phase, we achieve the ViMMRC dataset.

\textbf{Validation:} During this phase, primary-school teachers review the multiple-choice questions, their answer options, and their correct options again to ensure there are no mistakes. Finally, we obtain a highly-qualified dataset for research purposes for the computer multiple-choice reading comprehension mechanism. Table \ref{tab:examples} demonstrates some of the examples of Vietnamese multiple-choice MRC questions. In the following section, we analyze the characteristics of the dataset.

\begin{otherlanguage*}{vietnamese}
\begin{table}[ht]
\centering
\caption{Several examples of multiple-choice reading comprehension of Vietnamese texts are taken from our dataset. Each question has four answer options and there is only one correct answer in them. Besides, these Vietnamese examples are translated into English.}
\label{tab:examples}

\renewcommand{\arraystretch}{1}
\setlength\arrayrulewidth{1pt}
\begin{tabular}{lp{6cm}} 
\hline
Reading Text & \begin{tabular}[c]{p{6cm}}
\textbf{Vietnamese}: Ngay giữa sân trường, sừng sững một cây bàng. Mùa đông, cây vươn dài những cành khẳng khiu, trụi lá. Xuân sang, cành trên cành dưới chi chít những lộc non mơn mởn. Hè về, những tán lá xanh um che mát một khoảng sân trường. Thu đến, từng chùm quả chín vàng trong kẽ lá.\\
\textit{(\textbf{English translation}: In the middle of the schoolyard stood a towering tropical almond tree. In winter, the tree stretches out its slender, leafless branches. As spring arrives, its branches on the branches below are spangled with young buds. Summer approaches and its green foliage shades the yard. Autumn comes, revealing bunches of gold ripen fruits dangling in its leaves.)}\end{tabular}  \\ 
\hline
Question  & \begin{tabular}[c]{p{6cm}}Cây bàng được trồng ở đâu? \textit{(Where is the tropical almond tree planted?)}
\\A. Ngay giữa sân trường. \textit{(In the middle of the schoolyard.)} 
\\B. Trồng ở ngoài đường. \textit{(Planted on the street.)} 
\\C. Gần sông. \textit{(Near the river.)}
\\D. Dưới mái hiên trường. \textit{(Under the porch.)}
\end{tabular}                       \\ 

Answer    &  \begin{tabular}[c]{p{6cm}}  A  \end{tabular}
\\
\hline
Question  & \begin{tabular}[c]{p{6cm}}Những bộ phận nào của cây được nhắc đến trong bài đọc? \textit{(Which parts of the tree are mentioned?)}
\\A. Cành và lá. \textit{(Branches and leaves.)} 
\\B. Lá và quả. \textit{(Leaves and fruit.) }
\\C. Cành, lá, lộc, tán lá và quả. \textit{(Branches, leaves, buds, foliage and fruit.)}
\\D. Lộc, quả và tán cây. \textit{(Buds, fruit and foliage.)}
\end{tabular}                       \\ 

Answer    &   \begin{tabular}[c]{p{6cm}}  C  \end{tabular}
\\
\hline
\end{tabular}
\end{table}

\end{otherlanguage*}

\subsection{Dataset analysis}

\begin{table}[H]
\centering
\caption{Statistics about the training, development and test sets according to different aspects. The lengths are measured in words.}
\label{tab:datasetstatistics}
\setlength\arrayrulewidth{1pt}
\begin{tabular}{p{3.7cm}rrrr}
\hline
\multicolumn{1}{c}{} & \multicolumn{1}{c}{Train} & \multicolumn{1}{c}{Dev} & \multicolumn{1}{c}{Test} & \multicolumn{1}{c}{All} \\ \hline
Number of reading texts        & 292            & 42           & 83            & 417          \\ 
Number of questions       & 1,975          & 294          & 514           & 2,783        \\
Average text length            & 223.7         & 230.1       & 247.3        & 229.0       \\
Average question length           & 12.3          & 13.3        & 13.0         & 12.5        \\
Average answer option length             & 7.5           & 7.4         & 7.6          & 7.5         \\
Average correct answer length             & 8.7           & 8.4         & 8.9          & 8.7         \\
Vocabulary size           & 8,422          & 2,878        & 4,502         & 10,099       \\ \hline
\end{tabular}
\end{table}

We randomly divide our dataset into train, development, and test sets of 292 (70\%), 42 (10\%), and 83 (20\%) texts, respectively. The statistics of the training, development and test sets are summarized in Table \ref{tab:datasetstatistics}. 
In the table, the number of questions, the average words of reading texts, questions, answer options, correct answers, and vocabulary sizes are also listed.

\begin{table}[H]
\centering
\caption{Statistics of our dataset ViMMRC.}
\label{tab:2}
\setlength\arrayrulewidth{1pt}
\begin{tabular}{p{1.8cm}rrrrrr}
\hline
\multicolumn{1}{c}{Grade} & \multicolumn{1}{c}{1} & \multicolumn{1}{c}{2} & \multicolumn{1}{c}{3} & \multicolumn{1}{c}{4} & \multicolumn{1}{c}{5} & \multicolumn{1}{c}{All} \\ 
\hline
Number of texts  & 10  & 70 & 188  & 99 & 120   &417    \\ 
Vocabulary size (words) & 595  & 3,325 & 4,666   & 5,006 & 5,702  &10,099 \\
Number of questions & 60 & 514 & 759 & 709 & 741 & 2,783 \\ \hline
\end{tabular}
\end{table}

In this section, we present analysis of our dataset from different aspects. Table \ref{tab:2} shows statistics of our dataset with different grades. Vocabulary size, text length, question length, answer option length, and correct answer length are calculated in words. We used the word segmentation pyvi\footnote{Vietnamese word segmentation tool: https://pypi.org/project/pyvi/}. We found that the number of reading texts for the $1^{st}$ grade is small, which is obvious because the $1^{st}$ grade focuses on developing basic language skills rather than reading comprehension skill. We can observe that the vocabulary size increases as the grade increases. It can be inferred that the vocabulary sizes are correlated with the difficulty level of the reading comprehension task.

The types of reasoning required to solve the multiple-choice machine reading comprehension (MMRC) task directly influence the performance of MMRC models. In this paper, we classify the questions in our dataset following the same reasoning types as used in the analysis of the well-known dataset RACE \cite{lai2017}. These types are shown as follows, in ascending order of the difficulty level:

\begin{itemize}
\item \textbf{Word matching (WM)}: Important tokens in the question exactly match tokens in the reading text. Thus, it is easy to use a keyword search algorithm for finding the correct answer of this question based on the reading text.
\item \textbf{Paraphrasing (PP)}: The question is paraphrased from a single sentence in the reading text. In particular, we may use synonymy and world knowledge to create the question.
\item \textbf{Single-sentence reasoning (SSR)}: The answer is inferred from a single sentence in the reading text. Such answers could be created by extracting incomplete information or conceptual overlap.
\item \textbf{Multi-sentence reasoning (MSR)}: The answer is inferred from multiple sentences in the reading text by information synthesis techniques.
\item \textbf{Ambiguous or insufficient (AoI)}: The question has many answers or answers are not found in the reading text.
\end{itemize}

We manually annotate all questions in our dataset according to these types. Examples and percentages of these type are listed in Table \ref{tab:reasoning}. It can be seen from the table that single-sentence reasoning and ambiguous-or-insufficient make up the lowest proportions in our dataset (7.35\% for single-sentence reasoning and 6.12\% for ambiguous-or-insufficient). Meanwhile, word matching and multiple-sentence reasoning types account for the largest percentage, at 25.85\% and 36.73\% respectively. This demonstrates that ViMMRC is a challenging dataset for evaluating reading comprehension models for the Vietnamese language.

\subsection{Comparison with the MCTest dataset}
In this section, we compare our dataset with the MCTest dataset. The size of the MCTest dataset is approximately the same as our dataset. Table \ref{comparison} shows differences between our dataset and the MCTest dataset. As can be seen from the table, although the number of reading texts in our dataset is less than that of the MCTest dataset, the number of questions of our dataset is greater. Besides, the average numbers of words per reading text, per question and per answer in our dataset are also higher than those of the MCTest dataset.

\begin{table}[H]
\centering
\caption{Comparison between our dataset and the MCTest dataset.}
\label{comparison}
\setlength\arrayrulewidth{1pt}
\begin{tabular}{lrrrrr}
\hline
\multicolumn{1}{c}{\multirow{2}{*}{}} & \multicolumn{1}{c}{\multirow{2}{*}{\#Text}} & \multicolumn{1}{l}{\multirow{2}{*}{\#Question}} & \multicolumn{3}{c}{Average words per:}  \\ 
\cline{4-6} 
\multicolumn{1}{c}{}                                  & \multicolumn{1}{c}{}                                 & \multicolumn{1}{l}{}                                     & \multicolumn{1}{c}{Text} & \multicolumn{1}{c}{Question} & \multicolumn{1}{c}{Answer} \\ \hline
MCTest (160) & 160 & 640 & 204 & 8.0 & 3.4    \\
MCTest (500) & 500 & 2,000 & 212 & 7.7 & 3.4  \\
MCTest (560) & 660  & 2,640 & 210  & 7.8 & 3.4  \\
Our dataset & 417   & 2,783 & 229  & 12.5  & 7.5 \\ 
\hline
\end{tabular}
\end{table}

\begin{figure*}[]
\centering
\includegraphics[scale=0.85]{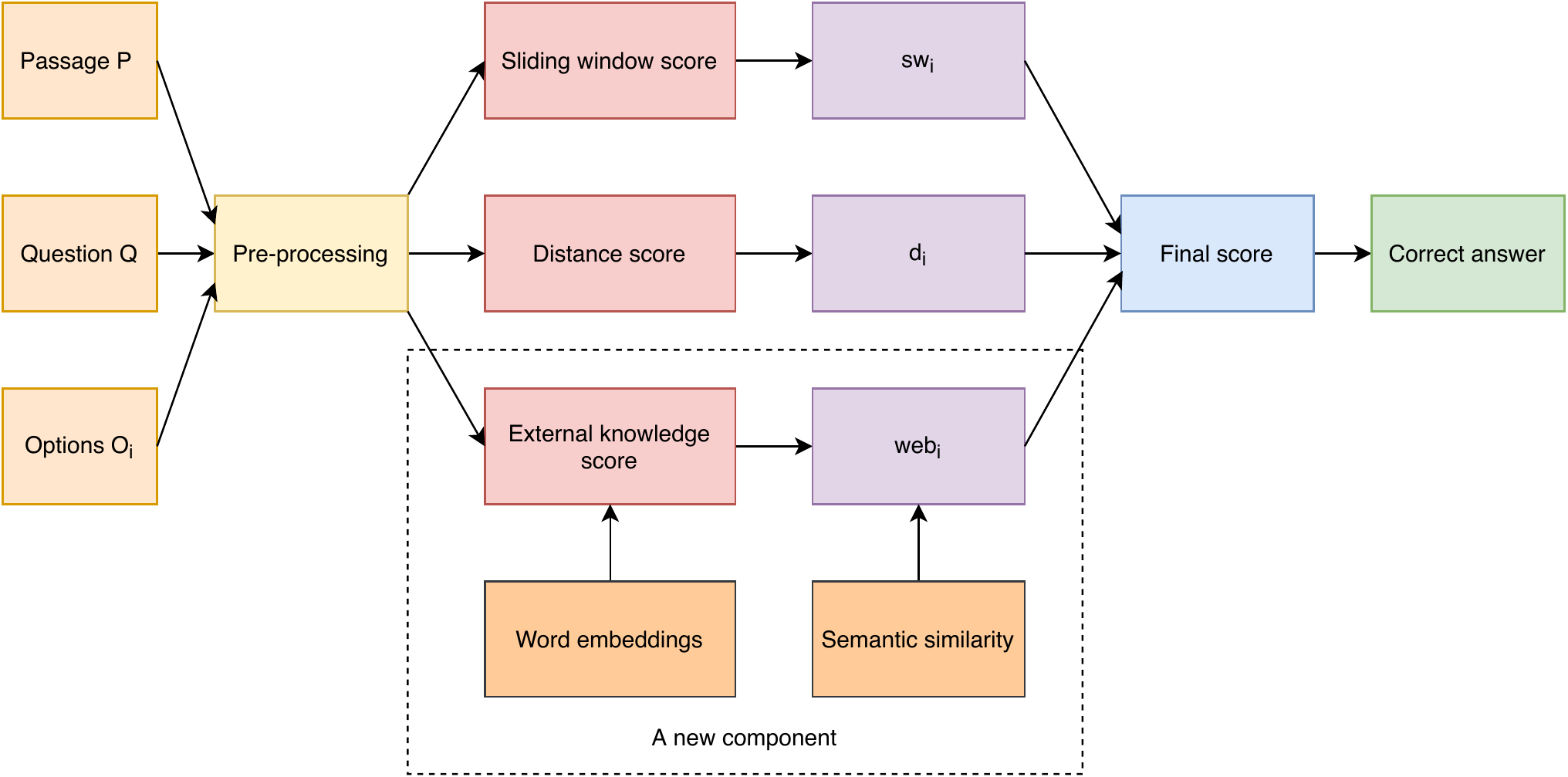}
\caption{System overview of our proposed method.}\label{fig:systemoverview}
\end{figure*}

\section{\textbf{Methodology}}
\label{method}

In this section, we introduce our proposed approach for the Vietnamese multiple-choice machine reading comprehension corpus. Because deep learning methods require a large dataset, so we only focus on the development of lexical-based methods on our dataset. Fig. \ref{fig:systemoverview} presents our proposed model by integrating semantic similarity and external knowledge sources into the lexical-based approach. This method is briefly described as follows. First of all, we pre-process the texts. Next, we calculate the sliding windows scores, the distance scores, and the external knowledge scores, respectively. Last, we combine those three scores for calculating the final score. The final score is used for predicting the correct answer in our approach. We implement this system through the algorithms described in detail as follows.

\subsection{Pre-processing techniques}
Pre-processing techniques play an important role in many applications of NLP. These techniques help to get rid of meaningless and confusing words, so we clean this data by following the steps shown in Algorithm \ref{al:prep1} and Algorithm \ref{al:prep2}. There are many techniques in natural language processing which are implemented in the pre-processing phase. In particular, Algorithm \ref{al:prep1} pre-processes for a sentence, applied to sentence processing in the reading text, questions and answer options.

\begin{algorithm}[H]
\caption{- \textbf{Pre-processing a raw Vietnamese sentence $S$}}
\label{al:prep1}

\begin{flushleft}
\textbf{Input:} A raw Vietnamese sentence $S$. \\
\textbf{Output:} A list of Vietnamese words after pre-processing $L$.  
\end{flushleft}

\begin{algorithmic}
\Procedure{Pre-processing a Vietnamese sentence}{}
\State   $X$ = tokenizing $S$ into a list of tokens.
\State Removing punctuations in $X$.
\State Removing Vietnamese stop words in $X$.
\State $S'$ = converting $X$ into a lower-case sentence.
\State $L$ = segmenting $S'$ into a list of words by the Vietnamese word segmentation.
\State \textbf{return} $L$.
\EndProcedure
\end{algorithmic}

\end{algorithm}

In Algorithm \ref{al:prep1}, firstly we use the tokenizer  to break a sentence into a list of Vietnamese tokens $X$. In our work, this step performs in three steps, removing punctuation marks, stop words and noise words (short vowels) in the list $X$. After that, we convert the list $X$ into a lower-case sentence $S'$. Lastly, we use the Vietnamese word segmentation tool to parse the sentence $S'$ into a list of Vietnamese words $L$ which is the output of this algorithm. We also apply Algorithm \ref{al:prep1} to both questions and answer options. We use the tool pyvi\footnote{Vietnamese word segmentation tool: https://pypi.org/project/pyvi/} for word segmentation in this algorithm.

\begin{algorithm}[]
\caption{- \textbf{Pre-processing a Vietnamese reading text T}}\label{al:prep2}

\begin{flushleft}
\textbf{Input:} A Vietnamese reading text $T$. \\
\textbf{Output:} A pre-processed reading text $T'$.  
\end{flushleft}

\begin{algorithmic}
\Procedure {Pre-processing a Vietnamese reading text}{}
\State $L$ = splitting $T$ into a list of single sentences.
\For {i = 1 to len($L$)}
\State $L_i$ = Pre-processing for a raw Vietnamese sentence($L_i$).
\EndFor
\State $T'$ = a pre-processed reading text converted from the list $L$.
\State \textbf{return $T'$}.
\EndProcedure
\end{algorithmic}
\end{algorithm}

In Algorithm \ref{al:prep2}, first of all, we split an input reading text into a list of sentences $L$. Then, we run the Pre-processing function (see Algorithm \ref{al:prep1}) for each sentence on all items of the list $L$. The output of this algorithm is a pre-processed reading text $T'$ converted from the list $L$. Algorithm \ref{al:prep1} and Algorithm \ref{al:prep2} are implemented in reading texts and multiple-choice questions on MMRC models.

\subsection{Sliding window and distance scores}

\begin{algorithm}[]
\caption{- \textbf{Calculating the sliding-widow scores}
\label{alg:sw}}

\begin{flushleft}
\textbf{Input:} Reading text $T$,\:set of words in question $Q$,\:and set of words in answer options $O_{1..4}$.\\
\textbf{Output: }Returning the score of the best answer option.   
\end{flushleft}

\begin{algorithmic}

\Procedure{Calculating sliding-window scores}{}
\State{\texttt{$C(w) = Count(w, T)$}}
\State{Initialize a list $sw$ of sliding-window scores for answer options.}
\For{i = 1 to len(O)}
    \State{\texttt{$  S = O_i \cup Q$}}
    \State {\resizebox{.85\hsize}{!}{$sw_{i}=max_{j=1}^{|T|}\sum _{l=1}^{|S|}\left\{\begin{matrix} &log(1+\frac{1}{C(T_{j+l})}) &,if \: T_{j+l}  \in S  \\  &0  &,Otherwise\end{matrix}\right.$}}
    
\EndFor

\State \textbf{return} $sw$
\EndProcedure
\end{algorithmic}
\end{algorithm}

We present how to calculate sliding window scores (see Algorithm \ref{alg:sw}) and distance scores (see Algorithm \ref{alg:dsw}) in the original sliding window algorithm (SW), a lexical-based approach developed by \cite{richardson2013}. This approach matches a bag of words, constructed from a question Q and an answer option $O_i$, with a given reading text, and calculates a TF-IDF style matching score for each answer option. The two algorithms are important components in our proposed model. To understand this method, we start with formal definitions of Vietnamese multiple-choice reading comprehension task. Let $T$ denote the reading text, $Q$ denote the question text, O$_{1..4}$  denote the texts of four answer options. The aim of the task is to predict the correct one among four answer options O$_{1..4}$ with regard to the question Q and the given reading text $T$. We also attempt to adapt Vietnamese textual structures into the sliding window algorithm (SW) as first baseline models on our proposed dataset. The results of these models are presented in Section \ref{experiment}.

\begin{algorithm}
\caption{- \textbf{Calculating the distance scores}
\label{alg:dsw}}

\begin{flushleft}
\textbf{Input:} Text T,\: set of reading-text words TW,\: set of words in question Q,\:and set of words in answer options $O_{1..4}$.\\
\textbf{Output: }Returning the distance score for answer options of the question.   
\end{flushleft}

\begin{algorithmic}[H]
\Procedure{Calculating distance scores}{}
\State{Initialize a list $d$ of distance score for answer options.}
\For{i = 1 to len(O)}
    
    \State{\texttt{$SQ = Q \cap TW$}}
    \State{\texttt{$SO_i = O_i \cap TW$}}
    
    \If{$|SQ| = 0$ or $|SO_i| = 0$}

    \State{\texttt{$d_i = 1$}}
    \Else
    \State{$d_i = \frac{1}{|T|-1} \max_{q \in SQ, a \in SO_i} d(T, q, a)$}
    \EndIf
    \State{where $d(T, q, a)$ is the minimum number of words an occurrence of $q$ and an occurrence of $a$ in $T$, increase 1}
    
\EndFor
\State \textbf{return} $d$
\EndProcedure
\end{algorithmic}
\end{algorithm}

\subsection{External Knowledge Integration}

\begin{figure}[ht!]
\centering
\includegraphics[scale=0.5]{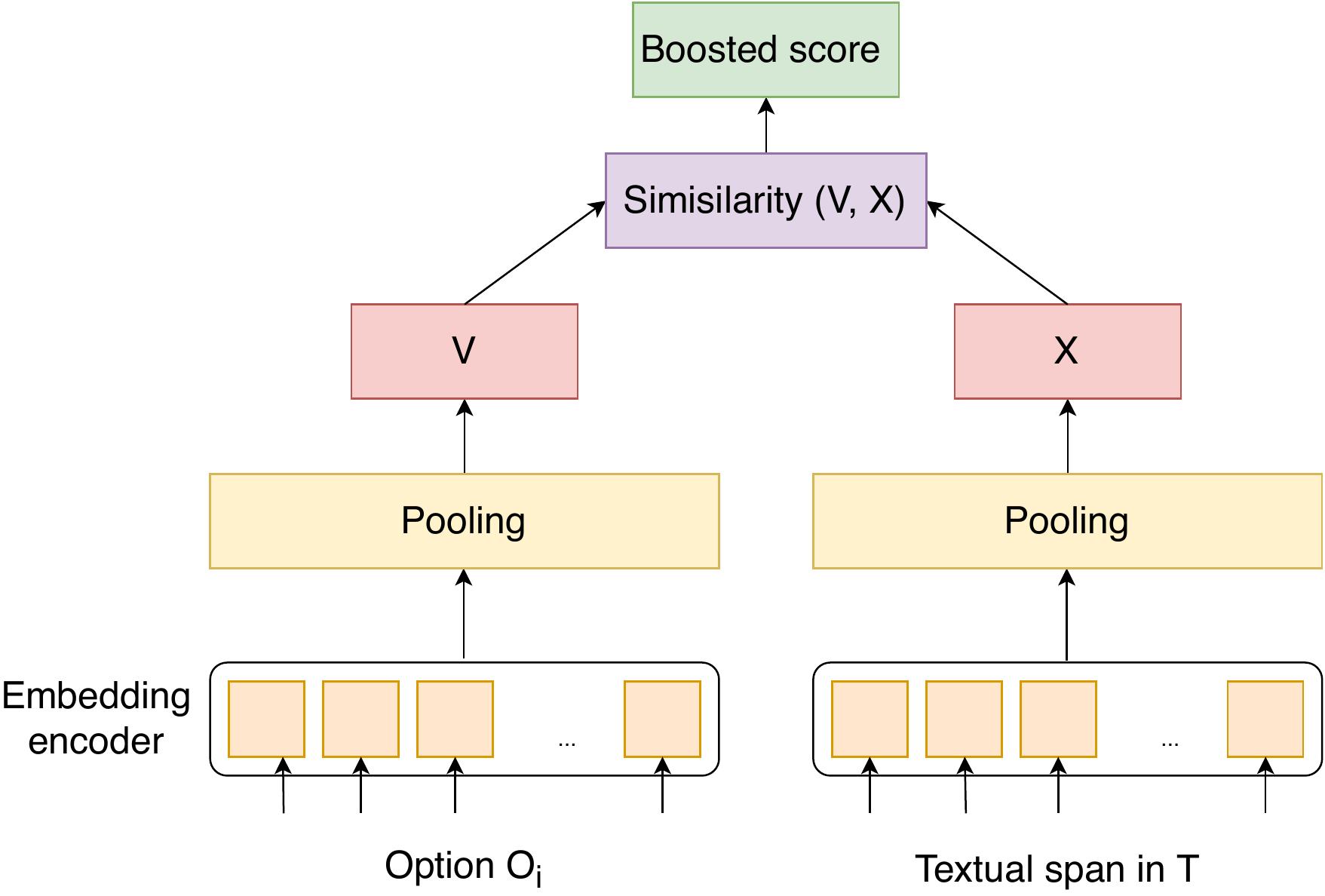}
\caption{Semantic similarity architecture estimating the boosted core between an answer option and a span in T. }\label{fig:semanticsimilarity}
\end{figure}

In addition to the lexical-based approach, we attach one more element to enrich world knowledge using semantic similarity and external knowledge sources like word embeddings. In particular, we add a boosted score (denoted by $web_{i}$) to the final score of each answer option. Algorithm \ref{al:prep5} presents how to calculate the boosted score. To understand Algorithm \ref{al:prep5}, we introduce two notations $V^T$ and $V^{O_i}$ to denote the ordered sets of words in the reading text $T$ and in the answer option O$_i$, respectively. We calculate $web_{i}$, the maximum cosine similarity between $V^{O_i}$ and span words X of the same length in $V^T$. $\overline{v}$ is the average of the word embeddings of the lexical units in $v$. Fig. \ref{fig:semanticsimilarity} shows semantic similarity architecture estimating the boosted core between an answer option and a span in T. The semantic similarity of the two vectors $V^{O_i}$ and $X$ is formulated as follows.

\begin{equation}
\begin{gathered}
{similarity(\overline{V^{O_i}},\overline{X})=cos(\overline{V^{O_i}},\overline{X})=\frac{\overline{V^{O_i}}.\overline{X}}{|\overline{V^{O_i}}|.|\overline{X}|}
}
\end{gathered}
\end{equation}

In this model, we use external knowledge sources as word embeddings. To explore the effectiveness of word embeddings, we evaluate the performance of our proposed model on with several word embeddings including Word2vec \cite{mikolov2013}, Word2vec and Character2vec \cite{kim2015}, fastText \cite{bojanowski2016}, ELMo \cite{peter2018}, BERT \cite{devlin2018} and MULTI \cite{Vu2019}. In particular, we use pre-trained embeddings on Vietnamese Wikipedia proposed by \cite{Vu2019} for all experiments of our proposed method.

Based on the above algorithms, we can regard $sw_i$ as the  sliding window score (see Algorithm \ref{alg:sw}) and $d_i$ as the distance score (see Algorithm \ref{alg:dsw}) defined in the original sliding window approach. In addition, the final distance-based sliding window score of $O_i$ \cite{richardson2013} can be formulated as follows.

\begin{equation}
\begin{gathered}
argmax_{i=1}^{|O|} (sw_i - d_i)
\end{gathered}
\end{equation}

Because a large proportion of questions cannot be solved by lexical-based approaches, we also try to incorporate external sources as general world knowledge into our lexical-based method. We calculate the boosted score for answer options of the question $web_i$ presented by Algorithm \ref{al:prep5}. To make the final answer option prediction, our lexical-based method combines the sliding widow score $sw_i$, 
the distance score $d_i$, and the boosted score  $web_i$ (see Algorithm \ref{al:prep5}) can be formulated as follows.

\begin{equation}
    \begin{gathered}
    argmax_{i=1}^{|O|} (sw_i - d_i + web_{i})
    \end{gathered}
\end{equation}

 \begin{algorithm}[H]
\caption{\textbf{Calculating the boosted score for answer options of a question}}
\label{al:prep5}

\begin{flushleft}
\textbf{Input:} Reading text $T$,\: set of reading-text words $TW$,\: set of words in question $Q$,\:and set of words in answer options $O_{1..4}$. $V^T$ and $V^{O_i}$ to denote the ordered sets of words in the reading text $T$ and in the answer option O$_i$, respectively. $\overline{v}$ is the average of the word embeddings of the lexical units in $v$. Note that $T$, $Q$, and $O_{1..4}$ are pre-processed by Algorithm \ref{al:prep1} and Algorithm \ref{al:prep2}.\\
\textbf{Output: }Returning the boosted score for answer options of a question.
\end{flushleft}

\begin{algorithmic}
\Procedure{Calculating the boosted scores}{}
\State{Initialize a list $web$ of boosted score for answer options.}
\For{i = 1 to len(O)}

    \State{$web_i = \max_{j}^{|T|}similarity\left( \overline{V^{O_i}}, \overline{V^T_{j ... j+|V^{O_i}| - 1}}\right)$}
    
\EndFor
\State \textbf{return} $web$
\EndProcedure
\end{algorithmic}
\end{algorithm}




\section{Empirical evaluation}
\label{experiment}
In this section, we compare the performance results of our proposed model with baseline models, and humans on our dataset. 

\subsection{Baseline models}
We evaluate random, lexical-based approaches (Sliding Window and Distance-based Sliding Window \cite{richardson2013}) and a neural network based model (Co-match \cite{wang2018}) as baseline models. Sliding Window and Distance-based Sliding Window were used in the baseline methods of various datasets such as MCTest \cite{richardson2013}, SQuAD \cite{squadr2016} and RACE \cite{lai2017}. Co-match is a strong neural network based model in multiple choice machine reading comprehension. Co-match achieve the positive results on RACE \cite{lai2017} and also was chosen to be the first baseline models on COSMOS QA \cite{huang2019cosmos} and C3 \cite{sun2020investigating}. Despite the limited data size of our dataset, we verify to evaluate how well the Co-match method do and then analyze the need for increasing training data for neural network based models presented in Sub-section \ref{sizeeff}.

\subsection{Evaluation metric and experimental settings}

We use accuracy as the primary evaluation metric which is computed as follows:

\begin{equation}
\begin{gathered}
Accuracy = \frac{
  \text{Number of questions correctly answered}
}{
  \text{Total number of questions}
}
\end{gathered}
\end{equation}

In all experiments, we use the word segmentation tool pyvi\footnote{Vietnamese word segmentation tool: https://pypi.org/project/pyvi/} and six different pre-trained word embeddings proposed by \cite{Vu2019}. The training, development and test sets are divided as shown in Table \ref{tab:datasetstatistics}. Besides, we implement three methods such as Random, Sliding Window and Distance-based Sliding Window as baseline models on our dataset.

For the model Co-match, we fine-tune the model parameters suitable for the Vietnamese multiple-choice MRC. In particular, we use a mini-batch size of 32, and the hidden memory size of 10. The number of epochs is set to a number of 30. Adamax optimizer is used for optimization with a starting learning rate of 0.002. In this model, we also in turn test the same word embeddings used on our proposed model.

\subsection{Human performance}
We randomly take 100 questions from the test set and 100 questions from the development set. We conduct the tests on ten students. As a result, human performance reaches 91.20\% in accuracy on the development set and 91.10\% on the test set. These results are much higher than our best model. To overcome human performance is challenging to explore a new machine reading comprehension model suitable for this dataset in the future.

\subsection{Model performance}

We report the performances of the baseline models and our proposed model in Table \ref{tab:results}.
Sliding Window and Distance-based Sliding Window achieve different performances, 58.50\% and 60.55\%, on the development set but they have the same accuracy of 56.30\% on the test set. 
Our proposed method achieves the accuracies over 60\% on the test set and over 61\% on the development set. Specifically, this method with the ELMo word embedding achieves the highest results on both of the test and development sets, 65.99\% and 61.81\%, respectively. This proves that our proposed method is more effective than the baseline methods for the Vietnamese MMRC task at present with improvements of 5.45\% and 5.51\% on the development and test sets, respectively. However, these results are much lower than the human performance of 29.29\% on the test set. This is a great challenge in the study of Vietnamese multiple-choice machine reading comprehension.

\begin{table}[H]
\centering
\caption{Experimental results of different models with various pre-trained word embeddings on our dataset ViMMRC.}
\label{tab:results}
\setlength\arrayrulewidth{1pt}
\begin{tabular}{p{1.5cm}p{3.4cm}ll}
\hline
\multicolumn{1}{p{1.5cm}}{{\bf Model}}          &
\multicolumn{1}{l}{{\bf Method}}          & \multicolumn{1}{l}{{\bf Dev (\%)}} & \multicolumn{1}{l}{{\bf Test} (\%)} \\ 
\hline
\multirow{3}{*}{\makecell{Baselines}} & Random  & 24.49  & 24.80 \\
& Sliding Window (SW)  & 58.50  & 56.30             \\
& Distance-based SW & 60.54 & 56.30   \\
& Co-match + W2V & 43.97 & 41.49  \\
& Co-match + W2V-C2V & 43.77 & 43.87    \\
& Co-match + fastText   & 43.39 & 41.84  \\
& Co-match + Bert-base  & 42.61 & 43.88  \\
& Co-match + ELMo   & 45.58 &44.94    \\
& Co-match + Multi  & 43.00 & 43.23  \\ 
\hline
\multirow{2}{*}{\makecell{Our proposed \\ approach}} & Boosted score with W2V     & 61.91 & 60.04\\
& Boosted score with W2V-C2V  & 61.91  & 60.04 \\
& Boosted score with fastText  & 63.27  & 60.04 \\
& Boosted score with Bert-base & 63.27  & 61.24 \\
& Boosted score with ELMo & \textbf{65.99} & \textbf{61.81} \\
& Boosted score with Multi & 63.61  & 60.24 \\ 
\hline
\multicolumn{2}{c}{Human performance} &91.20 & 91.10 \\ 
\hline
\end{tabular}
\end{table}

Comparing the experimental results of the Co-match model with different word embeddings, we can see that ELMO only achieves the best accuracy of 45.58\% and 44.94\% on development and test sets. However, ELMO is still the best word embedding on both lexical-based and neural-based approaches. In addition, the best performance of the Co-match model on the test set is 16.87\% lower than that of our proposed model. It is also much lower than the human performance of 46.16\%. Because the data size is not large enough, we  evaluate  this model on  different sizes of training data in Sub-section \ref{sizeeff}, which helps us to make a decision whether to continue increasing the data size in future work.
\section{\textbf{Result Analysis}}
\label{error}
To gain insights into the best model (our proposed method with the ELMo embedding), we analyze the experimental results in terms of different aspects such as question length, reading-text level, reasoning type, and word embedding. Besides, we aim to evaluate how the size of our training set has an impact on the neural network-based method.

\subsection{Effects of the question length}
\begin{figure}[H]
    \centering
    \begin{tikzpicture}[scale=0.48]
    \begin{axis}[
        width  = 1*\textwidth,
        height = 8cm,
        major x tick style = transparent,
        ybar=1*\pgflinewidth,
        axis x line*=bottom,
        bar width=20pt,
        ymajorgrids = true,
        ylabel = {Accuracy (\%)},
        symbolic x coords={$\leq10$,11-15,16-20,21-25,$\geq26$},
        xtick = data,
        nodes near coords, 
        every node near coord/.append style={rotate=90, anchor=west},
        ymin=0,
        scaled y ticks = false,
        legend cell align=left,
        legend style={
                at={(1,1.05)},
                anchor=south east,
                column sep=1ex
        }
    ]
    
        \addplot[style={black,fill=seagreen,mark=none}]
             coordinates {($\leq10$,62.38) (11-15,58.68)(16-20,71.74)(21-25,58.82)($\geq26$,11.11)};
             
        \addplot[style={black,fill=lightblue,mark=none}]
            coordinates {($\leq10$,64.36) (11-15,64.46)(16-20,84.78)(21-25,64.71)($\geq26$,11.11)};
        \legend{Best baseline,Our proposed method}
    \end{axis}
    \end{tikzpicture}
    \caption{Performance comparison between the best baseline and our proposed method with different groups of the question length.}\label{lbl:performancecomparison}
\end{figure}
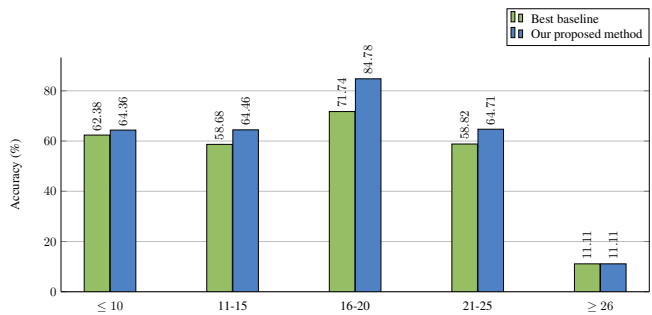

To verify whether the length of question is a reason for the poor performance of our best model, we measure the performances of the best model according to the question length. In particular, we divide the development set into five groups corresponding to the following question lengths: $\leq10$, $11-15$, $16-20$, $21-25$, and $\geq26$ words. The accuracies are analyzed and visualized in Fig. \ref{fig:length}. As can be seen from the figure, questions of the $16-20$ word length result in better performance than questions of other lengths. For short questions, our method predicts less effective. This may be because short questions contain less information beneficial to search for the correct answer. In particular, the performances on shorter questions (64.15\% for the $\leq10$-word questions and 65.18\% for $10-15$-word questions) are lower than the performances on longer questions which are over 66\% in accuracy. Fig. \ref{lbl:performancecomparison} shows performance comparison between the best baseline and our proposed method with different groups of the question length. The accuracy of our proposed model has an improvement on all question lengths (except the question lengths over 25 words), of which the three groups ($11-15$, $16-20$, $21-25$) have significant increases.

\begin{figure}[!ht]
    \centering
    
    \begin{minipage}{0.4\textwidth}
        \centering
        \begin{tikzpicture}[scale=0.7]
            \begin{axis}[
                ybar,
                bar width=19pt,
                symbolic x coords={$\leq10$,$11-15$,$16-20$,$21-25$,$\geq26$},
                xtick=data,
                ymin = 0,
                nodes near coords,
                nodes near coords align={vertical},
        	    ylabel near ticks,
        	    ylabel={Accuracy (\%)},
        	    xlabel near ticks,
        	    xlabel={Question length (words)},
            ]
            \addplot[black, fill=lightblue] coordinates {($\leq10$, 64.36) 
            ($11-15$,64.46) 
            ($16-20$,84.78) 
            ($21-25$, 64.71) 
            ($\geq26$, 11.11)};
            \end{axis}
        \end{tikzpicture}
    \end{minipage}
    \begin{minipage}{0.4\textwidth}
        \centering
        \setlength\arrayrulewidth{1pt}
        \begin{tabular}{cccc}\hline
          {\bf Question length} & {\bf \#Correct} & {\bf Total} & {\bf Acc. (\%)} \\ 
          \hline
            $\leq10$ & 65 & 101 & 64.36 \\
            $11-15$ & 78 & 121 & 64.46 \\
            $16-20$ & 39 & 46 & 84.78 \\
            $21-25$ & 12 & 17 & 64.71 \\
            $\geq26$ & 1 & 9 & 11.11 \\ \hline
          \end{tabular}
    \end{minipage}
    \caption{Analysis and visualization of the best model's result with different groups of the question length.}\label{fig:length}
\end{figure}
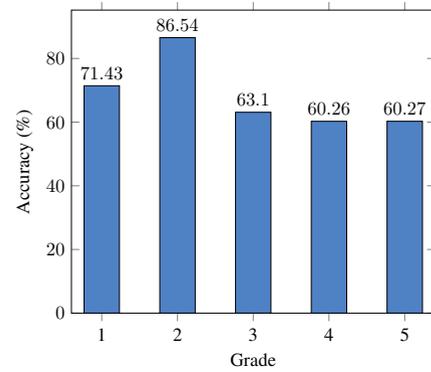

\subsection{Effects of the reading text level}
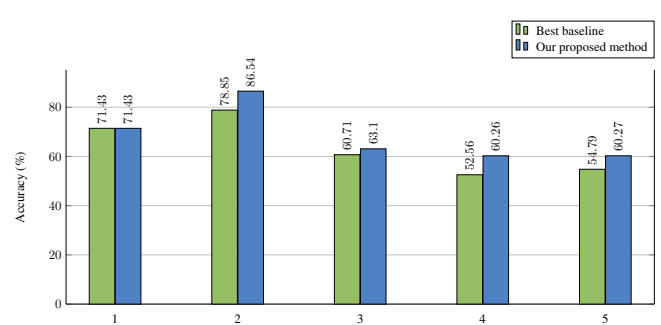
\begin{figure}[H]
    \centering
\begin{tikzpicture}[scale=0.48]
    \begin{axis}[
        width  = 1*\textwidth,
        height = 8cm,
        major x tick style = transparent,
        ybar=1*\pgflinewidth,
        axis x line*=bottom,
        bar width=20pt,
        ymajorgrids = true,
        ylabel = {Accuracy (\%)},
        symbolic x coords={1,2,3,4,5},
        xtick = data,
        nodes near coords, 
        every node near coord/.append style={rotate=90, anchor=west},
        ymin=0,
        scaled y ticks = false,
        legend cell align=left,
        legend style={
                at={(1,1.05)},
                anchor=south east,
                column sep=1ex
        }
    ]
    
        \addplot[style={black,fill=seagreen,mark=none}]
             coordinates {(1,71.43) (2,78.85)(3,60.71)(4,52.56)(5,54.79)};
             
        \addplot[style={black,fill=lightblue,mark=none}]
            coordinates {(1,71.43) (2,86.54)(3,63.10)(4,60.26)(5,60.27)};
    
        \legend{Best baseline,Our proposed method}
    \end{axis}
\end{tikzpicture}
    \caption{Performance comparison between the best baseline and our proposed method with different reading text levels.}\label{lbl:levelcomparison}
\end{figure}

Fig. \ref{fig:gradeanalysis} shows the accuracies of the best model according to different levels of reading text - the first to fifth grades. We can observe that the difficulty of the reading comprehension task increases together with the level of reading text. The system could answer questions of the $2^{nd}$ grade well, over 78\% in accuracy. It is more challenging to predict correct answers for questions of the $3^{rd}$ to $5^{th}$ grades (less than 68\%). The performance on $1^{st}$ grade questions is not as high as that on the $2^{nd}$ grade questions because the amount of questions of the $1^{st}$ grade is much fewer than those of other grades. Fig. \ref{lbl:levelcomparison} shows performance comparison between the best baseline and our proposed method with different reading text levels. The accuracy of our proposed model has an improvement on all reading text levels (except the first grade level), of which the three types (2-grade, 4-grade, and 5-grade) have significant increases. 

\begin{figure}[!htbp]
    \centering
    \begin{minipage}{0.4\textwidth}
        \centering
        \begin{tikzpicture}[scale=0.7]
            \begin{axis}[
                ybar,
                bar width=19pt,
                symbolic x coords={1,2,3,4,5},
                xtick=data,
                ymin = 0,
                nodes near coords,
                nodes near coords align={vertical},
        	    ylabel near ticks,
        	    ylabel={Accuracy (\%)},
        	    xlabel near ticks,
        	    xlabel={Grade},
            ]
            \addplot[black, fill=lightblue] coordinates {
            (1, 71.43) 
            (2,86.54) 
            (3,63.10) 
            (4, 60.26) 
            (5, 60.27)};
            \end{axis}
        \end{tikzpicture}
    \end{minipage}
    \begin{minipage}{0.4\textwidth}
        \centering
        \setlength\arrayrulewidth{1pt}
        \begin{tabular}{cccc}\hline
          {\bf Grade} & {\bf \#Correct} & {\bf Total} & {\bf Acc.(\%)} \\ \hline
            1 & 5 & 7 & 71.43 \\
            2 & 45 & 52 & 86.54 \\
            3 & 53 & 84 & 63.10 \\
            4 & 47 & 78 & 60.26 \\
            5 & 44 & 73 & 60.27 \\ \hline
          \end{tabular}
    \end{minipage}
    \caption{Analysis and visualization of the best model's result with different reading text levels.}
    \label{fig:gradeanalysis}
\end{figure}


\subsection{Effects of the reasoning type} 
\begin{figure}[H]
    \centering
\begin{tikzpicture}[scale=0.48]
    \begin{axis}[
        width  = 1*\textwidth,
        height = 8cm,
        major x tick style = transparent,
        ybar=1*\pgflinewidth,
        axis x line*=bottom,
        bar width=20pt,
        ymajorgrids = true,
        ylabel = {Accuracy (\%)},
        symbolic x coords={WM,PP,SSR,MSR,AoI},
        xtick = data,
        nodes near coords, 
        every node near coord/.append style={rotate=90, anchor=west},
        ymin=0,
        scaled y ticks = false,
        legend cell align=left,
        legend style={
                at={(1,1.05)},
                anchor=south east,
                column sep=1ex
        }
    ]
    
        \addplot[style={black,fill=seagreen,mark=none}]
             coordinates {(WM,84.21) (PP,70.73)(SSR,50.98)(MSR,48.15)(AoI,38.89)};
             
        \addplot[style={black,fill=lightblue,mark=none}]
            coordinates {(WM,92.11) (PP,82.93)(SSR,52.94)(MSR,50.00)(AoI,50.00)};
            
        \legend{Best baseline,Our proposed method}
    \end{axis}
\end{tikzpicture}
    \caption{Performance comparison between the best baseline and our proposed method with different reasoning types.}\label{lbl:reasoningcomparison}
\end{figure}
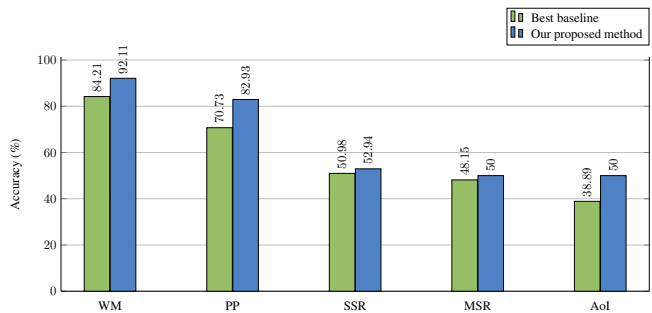

We also perform analysis to see how the reasoning types influence the best MMRC model. Fig. \ref{fig:reasoninganalysis} shows the analysis results. We found that the system determines answers more efficiently for the of the word matching and the paraphrasing reasoning types (WM and PP), 92.11\% and 82.93\% in accuracy, respectively. In contrast, complex forms of reasoning result in lower performances. They include single-sentence reasoning, multi-sentence reasoning, and ambiguous-or-insufficient. Fig. \ref{lbl:reasoningcomparison} shows performance comparison between the best baseline and our proposed method with different reasoning types. The accuracy of our proposed model has an improvement on all types of reasoning, of which the three types have significant increases: word matching (WM), paraphrasing (PP), and ambiguous or insufficient (AoI), while complex reasoning types (SSR and MSR) have slight improvements.

\begin{figure}[!h]
    \centering
    \begin{minipage}{0.4\textwidth}
        \centering
        \begin{tikzpicture}[scale=0.7]
            \begin{axis}[
                ybar,
                bar width=19pt,
                symbolic x coords={WM, PP, SSR, MSR, AoI},
                xtick=data,
                ymin = 0,
                nodes near coords,
                nodes near coords align={vertical},
        	    ylabel near ticks,
        	    ylabel={Accuracy (\%)},
        	    xlabel near ticks,
        	    xlabel={Reasoning types},
            ]
            \addplot[black, fill=lightblue] coordinates {
            (WM, 92.11) 
            (PP, 82.93) 
            (SSR, 52.94) 
            (MSR, 50.00) 
            (AoI, 50.00)};
            \end{axis}
        \end{tikzpicture}
    \end{minipage}
    \begin{minipage}{0.4\textwidth}
        \centering
        \setlength\arrayrulewidth{1pt}
        \begin{tabular}{cccc}\hline
          {\bf Reasoning type} & {\bf \#Correct} & {\bf Total} & {\bf Acc.(\%)} \\ \hline
            WM & 70 & 76 & 92.11 \\
            PP & 34 & 41 & 82.93 \\
            SSR & 27 & 51 & 52.94 \\
            MSR & 54 & 108 & 50.00 \\
            AoI & 9 & 18 & 50.00 \\ \hline
          \end{tabular}
    \end{minipage}
    
    \caption{Analysis and visualization of the best model's result with different types of reasoning.}
     \label{fig:reasoninganalysis}
\end{figure}

\subsection{Effects of the word embeddings} 
Table \ref{tab:results} shows the experimental results with external knowledge sources as pre-trained word embeddings. It can be seen that the results are influenced by the methods when combined with these word embeddings. In particular, our lexical-based approach achieves better results when using word embeddings, approximately 5\% higher. The experimental results show that ELMo is the best among the other word embeddings.

\begin{table*}[!htbp]
\centering
\caption{Statistics of the performance and improvement of our proposed model according to different lengths of question.}
\label{tbl:lenanalysis}
\begin{adjustbox}{width=0.8\textwidth}
\setlength\arrayrulewidth{1pt}
\begin{tabular}{ccccc}
\hline
\multicolumn{1}{l}{\multirow{2}{*}{\textbf{Question Length (words)}}} & \multicolumn{1}{l}{\multirow{2}{*}{\textbf{Ratio (\%)}}} & \multicolumn{2}{c}{\textbf{Accuracy (\%)}}                                                                                                               & \multicolumn{1}{l}{\multirow{2}{*}{\textbf{Improvement (\%)}}} \\ \cline{3-4}
\multicolumn{1}{l}{}                                          & \multicolumn{1}{c}{}                                & \multicolumn{1}{c}{\textbf{Best Baseline}} & \multicolumn{1}{c}{\textbf{\begin{tabular}[c]{@{}c@{}}Our Proposed Method \\ (Best Model)\end{tabular}}} & \multicolumn{1}{l}{}                                              \\ \hline
$\leq10$  & 36.05   & 62.38   & 64.36   & +1.98                                                               \\
$11-15$   & 38.10   & 58.68   & 64.46   & +5.78                                                               \\
$16-20$   & 16.67   & 71.74   & 84.78   & +13.04                                                               \\
$21-25$   & 6.12    & 58.82   & 64.71   & +5.89                                                              \\
$\geq26$  & 3.06    & 11.11   & 11.11   & 0.00                                                                  \\ \hline
\end{tabular}
\end{adjustbox}
\end{table*}

In addition, we conduct the detailed analysis of the effect of external knowledge integrated to our proposed model compared with the best baseline model according to different aspects such as the question length and reasoning type. In particular, Table \ref{tbl:lenanalysis} shows statistics of the performance and improvement of our proposed model according to different types of length. Our model improves the results of short questions ($\leq10$ 
words) with an increasing accuracy of 
1.98\% and average-length questions with an improvement of 5.78\% for $11-15$-word questions and the one of 13.04\% for $16-20$-word questions. For longer questions, this model does not improve its performance. However, this number is not significant because the number of long questions accounts for low percentage. Table \ref{tbl:reasoninganalysis} shows statistics of the performance and improvement of our proposed model according to different types of reasoning. We found that our proposed model is a right solution for three types of reasoning, word matching, paraphrasing and ambiguous or insufficient, increasing 7.90\%, 12.20\% and 11.11\% of the total number of solved questions, respectively. However, the number of questions of word matching and paraphrasing improved significantly because they account for a high proportion in the dataset.

\begin{table*}[!htbp]
\centering
\caption{Statistics of the performance and improvement of our proposed model according to different types of reasoning.}
\label{tbl:reasoninganalysis}
\begin{adjustbox}{width=0.8\textwidth}
\setlength\arrayrulewidth{1pt}
\begin{tabular}{ccccc}
\hline
\multicolumn{1}{l}{\multirow{2}{*}{\textbf{Reasoning Type}}} & \multicolumn{1}{c}{\multirow{2}{*}{\textbf{Ratio (\%)}}} & \multicolumn{2}{c}{\textbf{Accuracy (\%)}}                                                                                                               & \multicolumn{1}{l}{\multirow{2}{*}{\textbf{Improvement (\%)}}} \\ \cline{3-4}
\multicolumn{1}{l}{}                                         & \multicolumn{1}{c}{}                                & \multicolumn{1}{c}{\textbf{Best Baseline}} & \multicolumn{1}{c}{\textbf{\begin{tabular}[c]{@{}c@{}}Our Proposed Method \\ (Best Model)\end{tabular}}} & \multicolumn{1}{l}{}                                      \\ \hline
WM  & 25.85   & 84.21   & 92.11     & +7.90                                                      \\ 
PP  & 13.95   & 70.73   & 82.93     & +12.20                                                     \\
SSR & 17.35   & 50.98   & 52.94     & +1.96                                                      \\
MSR & 36.73   & 48.15   & 50.00     & +1.85                                                      \\
AoI & 6.12    & 38.89   & 50.00     & +11.11                                                     \\ \hline
\end{tabular}
\end{adjustbox}
\end{table*}

\subsection{Effects of the training data size}
\label{sizeeff}



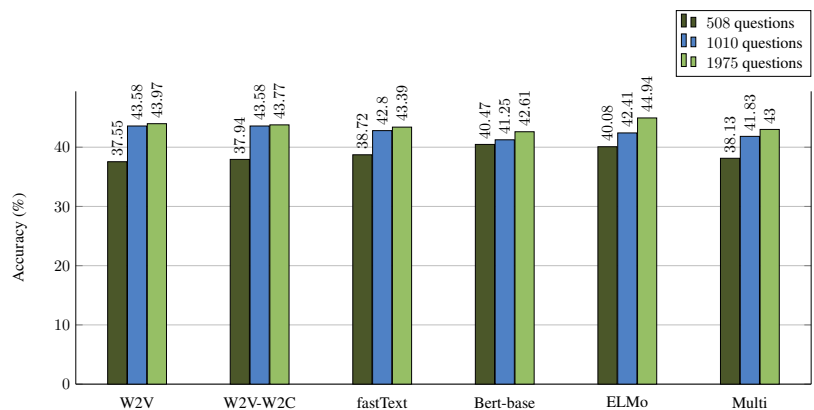
\begin{figure*}[!htbp]
    \centering
\begin{tikzpicture}[scale=0.6]
    \begin{axis}[
        width  = 1*\textwidth,
        height = 8cm,
        major x tick style = transparent,
        ybar=1*\pgflinewidth,
        axis x line*=bottom,
        bar width=12pt,
        ymajorgrids = true,
        ylabel = {Accuracy (\%)},
        symbolic x coords={W2V,W2V-W2C,fastText,Bert-base,ELMo,Multi},
        xtick = data,
        nodes near coords, 
        every node near coord/.append style={rotate=90, anchor=west},
        ymin=0,
        scaled y ticks = false,
        legend cell align=left,
        legend style={
                at={(1,1.05)},
                anchor=south east,
                column sep=1ex
        }
    ]
    
        \addplot[style={black,fill=olivegreen,mark=none}]
             coordinates {(W2V,37.55) (W2V-W2C,37.94)(fastText,38.72)(Bert-base,40.47)(ELMo,40.08)(Multi,38.13)};
             
        \addplot[style={black,fill=lightblue,mark=none}]
            coordinates {(W2V, 43.58) (W2V-W2C,43.58)(fastText,42.80)(Bert-base,41.25)(ELMo,42.41)(Multi,41.83)};


        \addplot[style={black,fill=seagreen,mark=none}]
             coordinates {(W2V,43.97) (W2V-W2C,43.77)(fastText,43.39)(Bert-base,42.61)(ELMo,44.94)(Multi,43.00)};

        \legend{$508$ questions,$1010$ questions,$1975$ questions}
    \end{axis}
\end{tikzpicture}

    \caption{Performances of Co-match with various word embeddings according to different sizes of training data.}\label{lbl:lncurve}
\end{figure*}

To verify whether the size of training data is a reason for the poor accuracy of the machine model, we evaluate Co-match \cite{wang2018} as a neural network-based model on different sizes of training data including $508$, $1010$ and $1975$ human-created questions. In this experiment, we implement Co-match with different pre-trained word embeddings \cite{Vu2019}. Experimental results (in accuracy) on the test set are presented in Fig. \ref{lbl:lncurve}. The figure shows that the model performance is improved when we increase the amount of the training data. These observations suggest that increasing training data size would improve the accuracy. This is also a future direction for addressing this problem.


\section{\textbf{Conclusion and future work}}
\label{conclusion}

In this paper, we propose the lexical-based approach utilizing semantic similarity and external knowledge sources and perform experiments to compare the performances between this method and baseline lexical-based and neural network based methods. The experimental results show that our proposed method is effective for Vietnamese multiple-choice reading comprehension. The best performance reaches 61.81\% in accuracy on our dataset. However, there is still a large gap between the human performance and the best model (a significant difference of 29.29\%). We also analyze the best models in different linguistic aspects to gain in-depth insights into the dataset. These analyses results illustrate that our corpus is a challenging dataset and need further studies. We also contribute this dataset for studies of the multiple-choice machine reading comprehension task for the Vietnamese language. This dataset includes 2,783 multiple-choice questions based on a set of 417 Vietnamese reading texts. This dataset encourages further advances in machine reading comprehension and guides the development of artificial intelligence for the Vietnamese language.

In future, we plan to increase the quantity and quality of the dataset in terms of the number of reading texts. The analysis results also suggest that we should focus on methods to improve the performance on long questions and difficult reasoning types.  When the dataset is large enough, we will further research on state-of-the-art methodologies such as deep neural networks and transfer learning to explore suitable models for Vietnamese multiple-choice MRC. In addition, we can use classification of the level of difficulty on multiple-questions to conduct experiments with curriculum learning \cite{liang2019new}.

\appendices
\section{Statistics of different reasoning types}
\label{app:reasoning}

Table \ref{tab:reasoning} shows the ratio of each reasoning type in the development set. Those types of reasoning have been described in Section \ref{data}. Besides, we have an example for each reasoning type and these Vietnamese examples are translated into English.

\begin{otherlanguage*}{vietnamese}
\begin{table*}[!ht]
    
    \caption{Statistics of different reasoning types in the development set of our dataset ViMMRC. The correct answer in each question is the answer option in bold. The options which are exact answers are italicized in the examples.} 
    
    \begin{tabular}{p{2cm}|p{13cm}|p{1cm}}
        \textbf{Reasoning Type} & \textbf{Example} & \textbf{Ratio (\%)} \\
        \hline
        Word Matching & \makecell*[{{p{13cm}}}]{
        \textbf{\textit{Reading text:}} \emph{Vừa sắp sách vở ra bàn}, Tường bỗng nghe \emph{có tiếng chuông điện thoại}. {(Just putting the books on the table, Tuong suddenly heard a phone ring.)}\\
        \textbf{\textit{Question}}: Việc gì đã xảy ra khi Tường \emph{vừa sắp sách vở ra bàn}? {(What happened when Tuong just put his books on the table?)}\\ 
        A. Mẹ nhờ Tường đi chợ. {(Mom asked Tuong to go to the market.)} \\
        \emph{B. Có tiếng chuông điện thoại}. {(There is a phone ringing.)} \\
        C. Bạn rủ Tường đi chơi. {(Tuong's friends invite him to go out.)}\\
        D. Nghe tiếng ai đó bên ngoài. {(Hearing someone's voice outside.)}\\} 
        & 25.85 \\
        \hline
        Paraphrasing & \makecell*[{{p{13cm}}}]{
        \textbf{\textit{Reading text:}} Tôi đang nắn nót viết từng chữ thì Cô-rét-ti chạm khuỷu tay vào tôi, làm cho cây bút nguệch ra một đường rất xấu. \\
        {(When I was sharpening the letters word for word, Coretti touched my elbows, making the pen scribble very badly.)}\\
        \textbf{\textit{Question:}} Khi nhân vật tôi đang nắn nót viết bài, chuyện gì đã xảy ra? {(When the character ``I'' was writing the lesson, what happened?)} \\ 
        A. Nhân vật tôi làm nguệch chữ đang viết của Cô-rét-ti. {(The character ``I`` made Coretti's written characters really ugly.)}\\
        B. Cô-rét-ti cãi cọ nhau vì một chữ viết nguệch. {(Coretti quarreled over a scribble.)} \\
        \emph{C. Cô-rét-ti chạm khuỷu tay làm tôi bị nguệch chữ. {(Coretti touched the elbows, making me scribble.)}}  \\
        D. Nhân vật tôi và Cô-rét-ti làm tranh nhau đồ dùng. {(The character ``I'' and Coretti competed together for getting something.)} \\
        }
        & 13.95 \\
        \hline
        Single-sentence Reasoning & \makecell*[{{p{13cm}}}]{
        \textbf{\textit{Reading text:}} Khi tiếng đàn, tiếng hát của A-ri-ôn vang lên, có một đàn cá heo đã bơi đến vây quanh tàu, say sưa thưởng thức tiếng hát của nghệ sĩ tài ba. \\ 
        {(As the sound of Arion's musical playing and singing started, a group of dolphins swam around the ship, passionately enjoying the singing of the talented artist.)}\\
        \textbf{\textit{Question:}} Điều kì lạ gì đã xảy ra khi nghệ sĩ A-ri-ôn cất tiếng hát giã biệt cuộc đời? ({What strange thing happened when the artist Arion sang goodbye to life? )}\\ 
        A. Đàn cá heo đã ăn thịt ông. {(The dolphins swallowed him.)}\\
        B. Đàn cá heo đã bỏ chạy đi mất. {(The dolphins ran away.)}\\
        C. Đàn cá heo đã nhấn chìm ông xuống biển. {(The dolphins drop him to the sea.)}\\
        \emph{D. Đàn cá heo đã bơi đến vây quanh tàu. {(The dolphins swam around the boat.)}} \\
        }
        & 7.35 \\
        \hline
        Multiple-sentence Reasoning & \makecell*[{{p{13cm}}}]{
        \textbf{\textit{Reading text:}} Chim đừng hót nữa, bà em ốm rồi, lặng cho bà ngủ. Bàn tay bé nhỏ, vẫy quạt thật đều. Ngấn nắng thiu thiu, đậu trên tường trắng. Căn nhà đã vắng. cốc chén nằm im. Đôi mắt lim dim, ngủ ngon bà nhé. \\
        {(Bird! Please don't sing, my grandma is sick, keep silent for her to sleep. Tiny hands are waving fans evenly. Sunlight stale parked on the white wall. The house is empty. The cup lies still. Eyes dim sleep. Sleep well, my grandma.)}\\
        \textbf{\textit{Question:}} Bạn nhỏ đang làm gì? {(What was the young boy doing ?)}\\ 
        A. Ngắm cây cối trong vườn. {(Viewing the trees in the garden.)}\\
        B. Nói chuyện với chim chích chòe. {(Talking with the warbler.)}\\
        C. Dọn dẹp nhà cửa. {(Cleaning his house.)}\\
        \emph{D. Quạt cho bà ngủ. {(Waving fans for his grandma's sleep.)}} \\
        }
        & 36.73 \\
        \hline
        Ambiguous or insufficient & \makecell*[{{p{13cm}}}]{
        \textbf{\textit{Reading text:}} Cậu bé nhìn bà, suy nghĩ một chút rồi thì thầm: những nếp nhăn, bà ạ!. \\
        {(The boy looked at his grandma, thought for a while and whispered: "The wrinkle, grandma!")}\\
        \textbf{\textit{Question:}} Câu trả lời cuối cùng của cậu bé muốn nói lên điều gì? {(What is the meaning of the boy's last answer?)}\\ 
        A. Cậu rất thích những người có nếp nhăn. {(The boy likes people with wrinkles very much.)}\\
        B. Cậu thấy những nếp nhăn rất đẹp. {(The boy thinks that wrinkles are very beautiful.)}\\
        \emph{C. Trong đôi mắt cậu, những nếp nhăn của bà rất đẹp và cậu rất yêu những nếp nhăn ấy. {(In the boy's eyes, wrinkles are very beautiful and he loves these wrinkles.)}}\\
        D. Trong đôi mắt cậu, hiện ra những vết nhăn của cô gái. {(In the boy's eyes, there are the girl's wrinkle.)}\\
        }
        & 6.12 \\
        \hline
    \end{tabular}
    
    \label{tab:reasoning}
\end{table*}
\end{otherlanguage*}

\section*{Acknowledgment}
We would like to appreciate the editors and anonymous reviewers for their helpful feedback.

\break
\bibliographystyle{IEEEtran}
\bibliography{reference}

\begin{thebibliography}{10}
\providecommand{\url}[1]{#1}
\csname url@samestyle\endcsname
\providecommand{\newblock}{\relax}
\providecommand{\bibinfo}[2]{#2}
\providecommand{\BIBentrySTDinterwordspacing}{\spaceskip=0pt\relax}
\providecommand{\BIBentryALTinterwordstretchfactor}{4}
\providecommand{\BIBentryALTinterwordspacing}{\spaceskip=\fontdimen2\font plus
\BIBentryALTinterwordstretchfactor\fontdimen3\font minus
  \fontdimen4\font\relax}
\providecommand{\BIBforeignlanguage}[2]{{%
\expandafter\ifx\csname l@#1\endcsname\relax
\typeout{** WARNING: IEEEtran.bst: No hyphenation pattern has been}%
\typeout{** loaded for the language `#1'. Using the pattern for}%
\typeout{** the default language instead.}%
\else
\language=\csname l@#1\endcsname
\fi
#2}}
\providecommand{\BIBdecl}{\relax}
\BIBdecl

\bibitem{hermann2015}
K.~M. Hermann, T.~Kocisky, E.~Grefenstette, L.~Espeholt, W.~Kay, M.~Suleyman,
  and P.~Blunsom, ``Teaching machines to read and comprehend,'' in
  \emph{Advances in neural information processing systems}, 2015, pp.
  1693--1701.

\bibitem{squadr2016}
P.~Rajpurkar, J.~Zhang, K.~Lopyrev, and P.~Liang, ``Squad: 100,000+ questions
  for machine comprehension of text,'' in \emph{Proceedings of the 2016
  Conference on Empirical Methods in Natural Language Processing}, 2016, pp.
  2383--2392.

\bibitem{choi2018}
E.~Choi, H.~He, M.~Iyyer, M.~Yatskar, W.~tau Yih, Y.~Choi, P.~Liang, and
  L.~Zettlemoyer, ``Quac: Question answering in context,'' in \emph{Proceedings
  of EMNLP}, 2018.

\bibitem{10.1145/3365679}
\BIBentryALTinterwordspacing
C.~Park, H.~Song, and C.~Lee, ``S3-net: Sru-based sentence and self-matching
  networks for machine reading comprehension,'' \emph{ACM Trans. Asian
  Low-Resour. Lang. Inf. Process.}, vol.~19, no.~3, Feb. 2020. [Online].
  Available: \url{https://doi.org/10.1145/3365679}
\BIBentrySTDinterwordspacing

\bibitem{Reddy2019}
S.~Reddy, D.~Chen, and C.~D. Manning, ``Coqa: A conversational question
  answering challenge,'' vol.~7, pp. 249--266, 2018.

\bibitem{qa1}
E.~M. Voorhees \emph{et~al.}, ``The trec-8 question answering track report,''
  in \emph{Trec}, vol.~99.\hskip 1em plus 0.5em minus 0.4em\relax Citeseer,
  1999, pp. 77--82.

\bibitem{qa2}
E.~M. Voorhees and H.~T. Dang, ``Overview of the trec 2002 question answering
  track,'' in \emph{TREC}, vol. 2003.\hskip 1em plus 0.5em minus 0.4em\relax
  Citeseer, 2003, pp. 54--68.

\bibitem{qa3}
P.~Nakov, D.~Hoogeveen, L.~M{\`a}rquez, A.~Moschitti, H.~Mubarak, T.~Baldwin,
  and K.~Verspoor, ``Semeval-2017 task 3: Community question answering,''
  \emph{arXiv preprint arXiv:1912.00730}, 2019.

\bibitem{10.1145/3359988}
\BIBentryALTinterwordspacing
D.~Gupta, A.~Ekbal, and P.~Bhattacharyya, ``A deep neural network framework for
  english hindi question answering,'' \emph{ACM Trans. Asian Low-Resour. Lang.
  Inf. Process.}, vol.~19, no.~2, Nov. 2019. [Online]. Available:
  \url{https://doi.org/10.1145/3359988}
\BIBentrySTDinterwordspacing

\bibitem{richardson2013}
M.~Richardson, C.~J. Burges, and E.~Renshaw, ``Mctest: A challenge dataset for
  the open-domain machine comprehension of text,'' in \emph{Proceedings of the
  2013 Conference on Empirical Methods in Natural Language Processing}, 2013,
  pp. 193--203.

\bibitem{mostafazadeh2016}
N.~Mostafazadeh, N.~Chambers, X.~He, D.~Parikh, D.~Batra, L.~Vanderwende,
  P.~Kohli, and J.~Allen, ``A corpus and cloze evaluation for deeper
  understanding of commonsense stories,'' in \emph{Proceedings of the
  NAACL-HLT}, 2016, pp. 839--849.

\bibitem{lai2017}
G.~Lai, Q.~Xie, H.~Liu, Y.~Yang, and E.~Hovy, ``Race: Largescale reading
  comprehension dataset from examinations,'' in \emph{Proceedings of the
  EMNLP}, 2017, pp. 785--794.

\bibitem{khashabi2018}
D.~Khashabi, S.~Chaturvedi, M.~Roth, S.~Upadhyay, and D.~Roth, ``Looking beyond
  the surface: A challenge set for reading comprehension over multiple
  sentences,'' in \emph{Proceedings of NAACL-HLT}, 2018, pp. 252--262.

\bibitem{ostermann2018}
S.~Ostermann, M.~Roth, A.~Modi, S.~Thater, and M.~Pinkal, ``Semeval-2018 task
  11: Machine comprehension using commonsense knowledge,'' in \emph{Proceedings
  of the SemEval}, 2018, pp. 747--757.

\bibitem{liu2019r}
S.~Liu, S.~Zhang, X.~Zhang, and H.~Wang, ``R-trans: Rnn transformer network for
  chinese machine reading comprehension,'' \emph{IEEE Access}, vol.~7, pp.
  27\,736--27\,745, 2019.

\bibitem{lui2020}
K.~Liu, X.~Liu, A.~Yang, J.~Liu, J.~Su, S.~Li, and Q.~She, ``A robust
  adversarial training approach to machine reading comprehension,'' in
  \emph{AAAI}, 2020.

\bibitem{yuan2020enhancing}
F.~Yuan, L.~Shou, X.~Bai, M.~Gong, Y.~Liang, N.~Duan, Y.~Fu, and D.~Jiang,
  ``Enhancing answer boundary detection for multilingual machine reading
  comprehension,'' \emph{arXiv preprint arXiv:2004.14069}, 2020.

\bibitem{lewis2019mlqa}
P.~Lewis, B.~O{\u{g}}uz, R.~Rinott, S.~Riedel, and H.~Schwenk, ``Mlqa:
  Evaluating cross-lingual extractive question answering,'' \emph{arXiv
  preprint arXiv:1910.07475}, 2019.

\bibitem{chen2016x}
J.~B. Danqi~Chen and C.~D. Manning, ``A thorough examination of the cnn/daily
  mail reading comprehension task,'' in \emph{Proceedings of the 54th Annual
  Meeting of the Association for Computational Linguistics}, 2016, p.
  2358–2367.

\bibitem{dhingra2016}
B.~Dhingra, H.~Liu, W.~W. Cohen, and R.~Salakhutdinov, ``Gated-attention
  readers for text comprehension,'' 2016.

\bibitem{zhu2018}
H.~Zhu, F.~Wei, B.~Qin, and T.~Liu, ``Hierarchical attention flow for
  multiple-choice reading comprehension,'' in \emph{Proceedings of the
  Thirty-Second AAAI Conference on Artificial Intelligence}, 2018.

\bibitem{luijcai2020}
X.~Liu, K.~Liu, X.~Li, J.~Su, Y.~Ge, B.~Wang, and J.~Luo, ``An iterative
  multi-source mutual knowledge transfer framework for machine reading
  comprehension,'' in \emph{IJCAI}, 2020.

\bibitem{bi2019incorporating}
B.~Bi, C.~Wu, M.~Yan, W.~Wang, J.~Xia, and C.~Li, ``Incorporating external
  knowledge into machine reading for generative question answering,''
  \emph{arXiv preprint arXiv:1909.02745}, 2019.

\bibitem{10.1145/3314936}
\BIBentryALTinterwordspacing
H.-L. Trieu, D.-V. Tran, A.~Ittoo, and L.-M. Nguyen, ``Leveraging additional
  resources for improving statistical machine translation on asian low-resource
  languages,'' \emph{ACM Trans. Asian Low-Resour. Lang. Inf. Process.},
  vol.~18, no.~3, Jun. 2019. [Online]. Available:
  \url{https://doi.org/10.1145/3314936}
\BIBentrySTDinterwordspacing

\bibitem{xu2019enhancing}
Y.~Xu, W.~Liu, G.~Chen, B.~Ren, S.~Zhang, S.~Gao, and J.~Guo, ``Enhancing
  machine reading comprehension with position information,'' \emph{IEEE
  Access}, vol.~7, pp. 141\,602--141\,611, 2019.

\bibitem{yang2020improved}
Y.~Yang, S.~Kang, and J.~Seo, ``Improved machine reading comprehension using
  data validation for weakly labeled data,'' \emph{IEEE Access}, vol.~8, pp.
  5667--5677, 2020.

\bibitem{sun2020investigating}
K.~Sun, D.~Yu, D.~Yu, and C.~Cardie, ``Investigating prior knowledge for
  challenging chinese machine reading comprehension,'' \emph{Transactions of
  the Association for Computational Linguistics}, vol.~8, pp. 141--155, 2020.

\bibitem{hill2016}
F.~Hill, A.~Bordes, S.~Chopra, and J.~Weston, ``The goldilocks principle:
  Reading children’s books with explicit memory representations,'' in
  \emph{Proceedings of the ICLR}, 2016.

\bibitem{trischler2016}
A.~Trischler, T.~Wang, X.~Yuan, J.~Harris, A.~Sordoni, P.~Bachman, and
  K.~Suleman, ``Newsqa: A machine comprehension dataset,'' in \emph{Proceedings
  of the 2nd Workshop on Representation Learning for NLP}, 2017, pp. 191--200.

\bibitem{chen2016}
Y.-H. Chen and J.~D. Choi, ``Character identification on multiparty
  conversation: Identifying mentions of characters in tv shows,'' in
  \emph{Proceedings of the 17th Annual Meeting of the Special Interest Group on
  Discourse and Dialogue}, 2016, pp. 90--100.

\bibitem{joshi2017}
M.~Joshi, E.~Choi, D.~S. Weld, and L.~Zettlemoyer, ``Triviaqa: A large scale
  distantly supervised challenge dataset for reading comprehension,'' in
  \emph{Proceedings of the 55th Annual Meeting of the Association for
  Computational Linguistics}, 2017, pp. 1601--1611.

\bibitem{welbl2018}
J.~Welbl, P.~Stenetorp, and S.~Riedel, ``Constructing datasets for multi-hop
  reading comprehension across documents,'' \emph{Transactions of the
  Association for Computational Linguistics}, vol.~6, pp. 287--302, 2018.

\bibitem{lee2018}
C.-H. Lee, S.-M. Wang, H.~Chang, and H.-Y. Lee, ``Odsqa: Open-domain spoken
  question answering dataset,'' in \emph{2018 IEEE Spoken Language Technology
  Workshop (SLT)}, 2018, pp. 949--956.

\bibitem{li2018}
C.-H. Li, S.-L. Wu, C.-L. Liu, and H.-y. Lee, ``Spoken squad: A study of
  mitigating the impact of speech recognition errors on listening
  comprehension,'' \emph{arXiv preprint arXiv:1804.00320}, 2018.

\bibitem{ma2018}
K.~Ma, T.~Jurczyk, and J.~D. Choi, ``Challenging reading comprehension on daily
  conversation: Passage completion on multiparty dialog,'' in \emph{Proceedings
  of NAACL-HLT}, 2018, pp. 2039--2048.

\bibitem{he2017dureader}
W.~He, K.~Liu, J.~Liu, Y.~Lyu, S.~Zhao, X.~Xiao, Y.~Liu, Y.~Wang, H.~Wu, Q.~She
  \emph{et~al.}, ``Dureader: a chinese machine reading comprehension dataset
  from real-world applications,'' \emph{arXiv preprint arXiv:1711.05073}, 2017.

\bibitem{shao2018drcd}
C.~C. Shao, T.~Liu, Y.~Lai, Y.~Tseng, and S.~Tsai, ``Drcd: a chinese machine
  reading comprehension dataset,'' \emph{arXiv preprint arXiv:1806.00920},
  2018.

\bibitem{cui2018span}
Y.~Cui, T.~Liu, W.~Che, L.~Xiao, Z.~Chen, W.~Ma, S.~Wang, and G.~Hu, ``A
  span-extraction dataset for chinese machine reading comprehension,'' in
  \emph{Proceedings of the 2019 Conference on Empirical Methods in Natural
  Language Processing and the 9th International Joint Conference on Natural
  Language Processing (EMNLP-IJCNLP)}, 2019, pp. 5886--5891.

\bibitem{nguyen2016}
\BIBentryALTinterwordspacing
T.~Nguyen, M.~Rosenberg, X.~Song, J.~Gao, S.~Tiwary, R.~Majumder, and L.~Deng,
  ``Ms marco: A human generated machine reading comprehension dataset,''
  November 2016. [Online]. Available:
  \url{https://www.microsoft.com/en-us/research/publication/ms-marco-human-generated-machine-reading-comprehension-dataset/}
\BIBentrySTDinterwordspacing

\bibitem{dunn2017}
M.~Dunn, L.~Sagun, M.~Higgins, V.~U. Guney, V.~Cirik, and K.~Cho, ``Searchqa: A
  new q\&a dataset augmented with context from a search engine,'' \emph{arXiv
  preprint arXiv:1704.05179}, 2017.

\bibitem{kocisk2018}
T.~Ko{\v{c}}isk{\`y}, J.~Schwarz, P.~Blunsom, C.~Dyer, K.~M. Hermann, G.~Melis,
  and E.~Grefenstette, ``The narrativeqa reading comprehension challenge,''
  \emph{Transactions of the Association for Computational Linguistics}, vol.~6,
  pp. 317--328, 2018.

\bibitem{huang2019cosmos}
L.~Huang, R.~Le~Bras, C.~Bhagavatula, and Y.~Choi, ``Cosmos qa: Machine reading
  comprehension with contextual commonsense reasoning,'' in \emph{Proceedings
  of the 2019 Conference on Empirical Methods in Natural Language Processing
  and the 9th International Joint Conference on Natural Language Processing
  (EMNLP-IJCNLP)}, 2019, pp. 2391--2401.

\bibitem{penas2014}
A.~Penas, Y.~Miyao, A.~Rodrigo, E.~H. Hovy, and N.~Kando, ``Overview of clef qa
  entrance exams task 2014,'' in \emph{Proceedings of the SemEval}, 2014, pp.
  1194--1200.

\bibitem{shibuki2014}
H.~Shibuki, K.~Sakamoto, Y.~Kano, T.~Mitamura, M.~Ishioroshi, K.~Y. Itakura,
  D.~Wang, T.~Mori, and N.~Kando, ``Overview of the ntcir11 qa-lab task,'' in
  \emph{Proceedings of NTCIR}, 2014.

\bibitem{tseng2016}
B.-H. Tseng, S.-S. Shen, H.-Y. Lee, and L.-S. Lee, ``Towards machine
  comprehension of spoken content: Initial toefl listening comprehension test
  by machine,'' in \emph{Proceedings of the Interspeech}, 2016.

\bibitem{clark2016}
P.~Clark, O.~Etzioni, T.~Khot, A.~Sabharwal, O.~Tafjord, P.~D. Turney, and
  D.~Khashabi, ``Combining retrieval, statistics, and inference to answer
  elementary science questions,'' in \emph{Proceedings of the AAAI}, 2016, pp.
  2580--2586.

\bibitem{wang2018}
S.~Wang, M.~Yu, J.~Jiang, and S.~Chang, ``A co-matching model for multi-choice
  reading comprehension,'' in \emph{Proceedings of the 56th Annual Meeting of
  the Association for Computational Linguistics}, 2018, pp. 746--751.

\bibitem{radford2019}
A.~Radford, K.~Narasimhan, T.~Salimans, and I.~Sutskever, ``Improving language
  understanding by generative pre-training,'' \emph{URL https://s3-us-west-2.
  amazonaws. com/openai-assets/researchcovers/languageunsupervised/language
  understanding paper. pdf}, 2018.

\bibitem{devlin2018}
J.~Devlin, M.-W. Chang, K.~Lee, and K.~Toutanova, ``Bert: Pre-training of deep
  bidirectional transformers for language understanding,'' 2018.

\bibitem{nguyen2014treebank}
D.~Q. Nguyen, D.~Q. Nguyen, S.~B. Pham, P.-T. Nguyen, and M.~Le~Nguyen, ``From
  treebank conversion to automatic dependency parsing for vietnamese,'' in
  \emph{International Conference on Applications of Natural Language to Data
  Bases/Information Systems}.\hskip 1em plus 0.5em minus 0.4em\relax Springer,
  2014, pp. 196--207.

\bibitem{nguyen2016x}
K.~V. Nguyen and N.~L.-T. Nguyen, ``Vietnamese transition-based dependency
  parsing with supertag features,'' in \emph{2016 Eighth International
  Conference on Knowledge and Systems Engineering (KSE)}.\hskip 1em plus 0.5em
  minus 0.4em\relax IEEE, 2016, pp. 175--180.

\bibitem{nguyen2018lstm}
B.~D. Nguyen, K.~Van~Nguyen, and N.~L.-T. Nguyen, ``Lstm easy-first dependency
  parsing with pre-trained word embeddings and character-level word embeddings
  in vietnamese,'' in \emph{2018 10th International Conference on Knowledge and
  Systems Engineering (KSE)}.\hskip 1em plus 0.5em minus 0.4em\relax IEEE,
  2018, pp. 187--192.

\bibitem{datpos}
D.~Q. Nguyen, D.~Q. Nguyen, D.~D. Pham, and S.~B. Pham, ``{RDRPOST}agger: A
  ripple down rules-based part-of-speech tagger,'' \emph{In: Proceedings of the
  Demonstrations at the 14th Conference of the {E}uropean Chapter of the
  Association for Computational Linguistics}, pp. 17--20, Apr. 2014.

\bibitem{bachpos}
N.~X. Bach, N.~D. Linh, and T.~M. Phuong, ``An empirical study on {POS} tagging
  for vietnamese social media text,'' \emph{Computer Speech \& Language},
  vol.~50, pp. 1--15, 2018.

\bibitem{10.1145/1316457.1316460}
\BIBentryALTinterwordspacing
P.~T.~X. Thao, T.~Q. Tri, D.~Dien, and N.~Collier, ``Named entity recognition
  in vietnamese using classifier voting,'' \emph{ACM Transactions on Asian
  Language Information Processing}, vol.~6, no.~4, Dec. 2008. [Online].
  Available: \url{https://doi.org/10.1145/1316457.1316460}
\BIBentrySTDinterwordspacing

\bibitem{10.1145/2990191}
\BIBentryALTinterwordspacing
L.~H.~B. Nguyen, D.~Dinh, and P.~Tran, ``An approach to construct a named
  entity annotated english-vietnamese bilingual corpus,'' \emph{ACM Trans.
  Asian Low-Resour. Lang. Inf. Process.}, vol.~16, no.~2, Oct. 2016. [Online].
  Available: \url{https://doi.org/10.1145/2990191}
\BIBentrySTDinterwordspacing

\bibitem{nguyen2019error}
B.~A. Nguyen, K.~Van~Nguyen, and N.~L.-T. Nguyen, ``Error analysis for
  vietnamese named entity recognition on deep neural network models,''
  \emph{arXiv preprint arXiv:1911.07228}, 2019.

\bibitem{van2018uit}
K.~V. Nguyen, V.~D. Nguyen, P.~X. Nguyen, T.~T. Truong, and N.~L.-T. Nguyen,
  ``Uit-vfsc: Vietnamese students’ feedback corpus for sentiment analysis,''
  in \emph{2018 10th International Conference on Knowledge and Systems
  Engineering (KSE)}.\hskip 1em plus 0.5em minus 0.4em\relax IEEE, 2018, pp.
  19--24.

\bibitem{nguyen2018deep}
P.~X. Nguyen, T.~T. Hong, K.~Van~Nguyen, and N.~L.-T. Nguyen, ``Deep learning
  versus traditional classifiers on vietnamese students’ feedback corpus,''
  in \emph{2018 5th NAFOSTED Conference on Information and Computer Science
  (NICS)}.\hskip 1em plus 0.5em minus 0.4em\relax IEEE, 2018, pp. 75--80.

\bibitem{van2018transformation}
T.~V. Dang, V.~D. Nguyen, K.~V. Nguyen, and N.~L.-T. Nguyen, ``A transformation
  method for aspect-based sentiment analysis,'' \emph{Journal of Computer
  Science and Cybernetics}, vol.~34, no.~4, pp. 323--333, 2018.

\bibitem{Nguyen_2009}
D.~Q. Nguyen, D.~Q. Nguyen, and S.~B. Pham, ``A vietnamese question answering
  system,'' in \emph{2009 International Conference on Knowledge and Systems
  Engineering}.\hskip 1em plus 0.5em minus 0.4em\relax IEEE, 2009.

\bibitem{van2016improving}
V.-T. Nguyen and A.-C. Le, ``Improving question classification by feature
  extraction and selection,'' \emph{Indian Journal of Science and Technology},
  vol.~9, no.~17, pp. 1--8, 2016.

\bibitem{le2018factoid}
P.~H. Le and D.-T. Bui, ``A factoid question answering system for vietnamese,''
  in \emph{Companion Proceedings of the The Web Conference 2018}.\hskip 1em
  plus 0.5em minus 0.4em\relax International World Wide Web Conferences
  Steering Committee, 2018, pp. 1049--1055.

\bibitem{10.1145/3182622}
\BIBentryALTinterwordspacing
D.~Huang, J.~Pei, C.~Zhang, K.~Huang, and J.~Ma, ``Incorporating prior
  knowledge into word embedding for chinese word similarity measurement,''
  \emph{ACM Trans. Asian Low-Resour. Lang. Inf. Process.}, vol.~17, no.~3, Apr.
  2018. [Online]. Available: \url{https://doi.org/10.1145/3182622}
\BIBentrySTDinterwordspacing

\bibitem{gupta2017continuous}
P.~Gupta, R.~E. Banchs, and P.~Rosso, ``Continuous space models for clir,''
  \emph{Information Processing \& Management}, vol.~53, no.~2, pp. 359--370,
  2017.

\bibitem{zhou2019text}
S.~Zhou, X.~Xu, Y.~Liu, R.~Chang, and Y.~Xiao, ``Text similarity measurement of
  semantic cognition based on word vector distance decentralization with
  clustering analysis,'' \emph{IEEE Access}, vol.~7, pp. 107\,247--107\,258,
  2019.

\bibitem{meng2013review}
L.~Meng, R.~Huang, and J.~Gu, ``A review of semantic similarity measures in
  wordnet,'' \emph{International Journal of Hybrid Information Technology},
  vol.~6, no.~1, pp. 1--12, 2013.

\bibitem{jiang2017wikipedia}
Y.~Jiang, W.~Bai, X.~Zhang, and J.~Hu, ``Wikipedia-based information content
  and semantic similarity computation,'' \emph{Information Processing \&
  Management}, vol.~53, no.~1, pp. 248--265, 2017.

\bibitem{rumelhart1986}
D.~E. Rumelhart, G.~E. Hinton, and R.~J. Williams, ``Learning representations
  by back-propagating errors,'' \emph{nature}, vol. 323, no. 6088, pp.
  533--536, 1986.

\bibitem{seo2017}
M.~Seo, A.~Kembhavi, A.~Farhadi, and H.~Hajishirzi, ``Bidirectional attention
  flow for machine comprehension,'' in \emph{Proceedings of ICLR 2017}, 2017.

\bibitem{hu2018}
M.~Hu, Y.~Peng, Z.~Huang, X.~Qiu, F.~Wei, and M.~Zhou, ``Reinforced mnemonic
  reader for machine reading comprehension,'' in \emph{Proceedings of the
  Twenty-Seventh International Joint Conference on Artificial Intelligence},
  2018, p. 4099–4106.

\bibitem{peter2018}
M.~E. Peters, M.~Neumann, M.~Iyyer, M.~Gardner, C.~Clark, K.~Lee, and
  L.~Zettlemoyer, ``Deep contextualized word representations,'' in
  \emph{Proceedings of NAACL-HLT 2018}, 2018, p. 2227–2237.

\bibitem{mikolov2013}
T.~Mikolov, K.~Chen, G.~Corrado, and J.~Dean, ``Efficient estimation of word
  representations in vector space,'' \emph{arXiv preprint arXiv:1301.3781},
  2013.

\bibitem{kim2015}
Y.~Kim, Y.~Jernite, D.~Sontag, and A.~M. Rush, ``Character-aware neural
  language models,'' in \emph{Thirtieth AAAI Conference on Artificial
  Intelligence}, 2016.

\bibitem{bojanowski2016}
P.~Bojanowski, E.~Grave, A.~Joulin, and T.~Mikolov, ``Enriching word vectors
  with subword information,'' \emph{Transactions of the Association for
  Computational Linguistics}, vol.~5, pp. 135--146, 2017.

\bibitem{Vu2019}
X.-S. Vu, T.~Vu, S.~N. Tran, and L.~Jiang, ``Etnlp: A visual-aided systematic
  approach to select pre-trained embeddings for a downstream task,'' in
  \emph{Proceedings of the International Conference Recent Advances in Natural
  Language Processing (RANLP)}, 2019.

\bibitem{liang2019new}
Y.~Liang, J.~Li, and J.~Yin, ``A new multi-choice reading comprehension dataset
  for curriculum learning,'' in \emph{Asian Conference on Machine Learning},
  2019, pp. 742--757.

\end{thebibliography}

\begin{IEEEbiography}[{\includegraphics[width=1in,height=1.25in,clip,keepaspectratio]{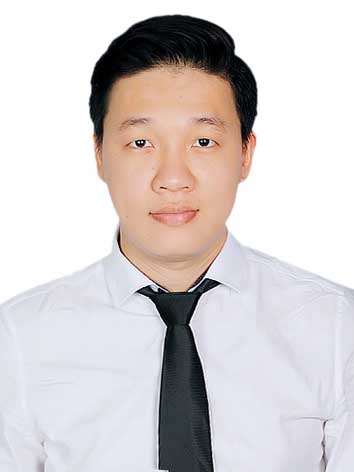}}]{Kiet Van Nguyen} is a lecturer of Faculty of Information Science and Engineering at University of Information Technology, Vietnam National University, Ho Chi Minh City, Vietnam. He obtained his B.S. and M.S. degrees from the University of Information Technology, Ho Chi Minh City, Vietnam in 2012 and 2017, respectively. His research interests include natural language processing, machine reading comprehension and deep learning.
\end{IEEEbiography}

\begin{IEEEbiography}[{\includegraphics[width=1in,height=1.25in,clip,keepaspectratio]{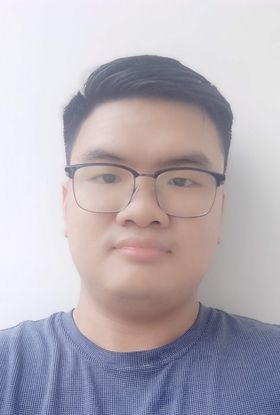}}]{Khiem Vinh Tran} is a junior student at University of Information Technology, Vietnam National University, Ho Chi Minh City, Vietnam. He just took part in the WNUT-2020 Task 2 and ranked the third place in this competition. His research interests include text processing, machine reading comprehension and sentiment analysis.
\end{IEEEbiography}

\begin{IEEEbiography}[{\includegraphics[width=1in,height=1.25in,clip,keepaspectratio]{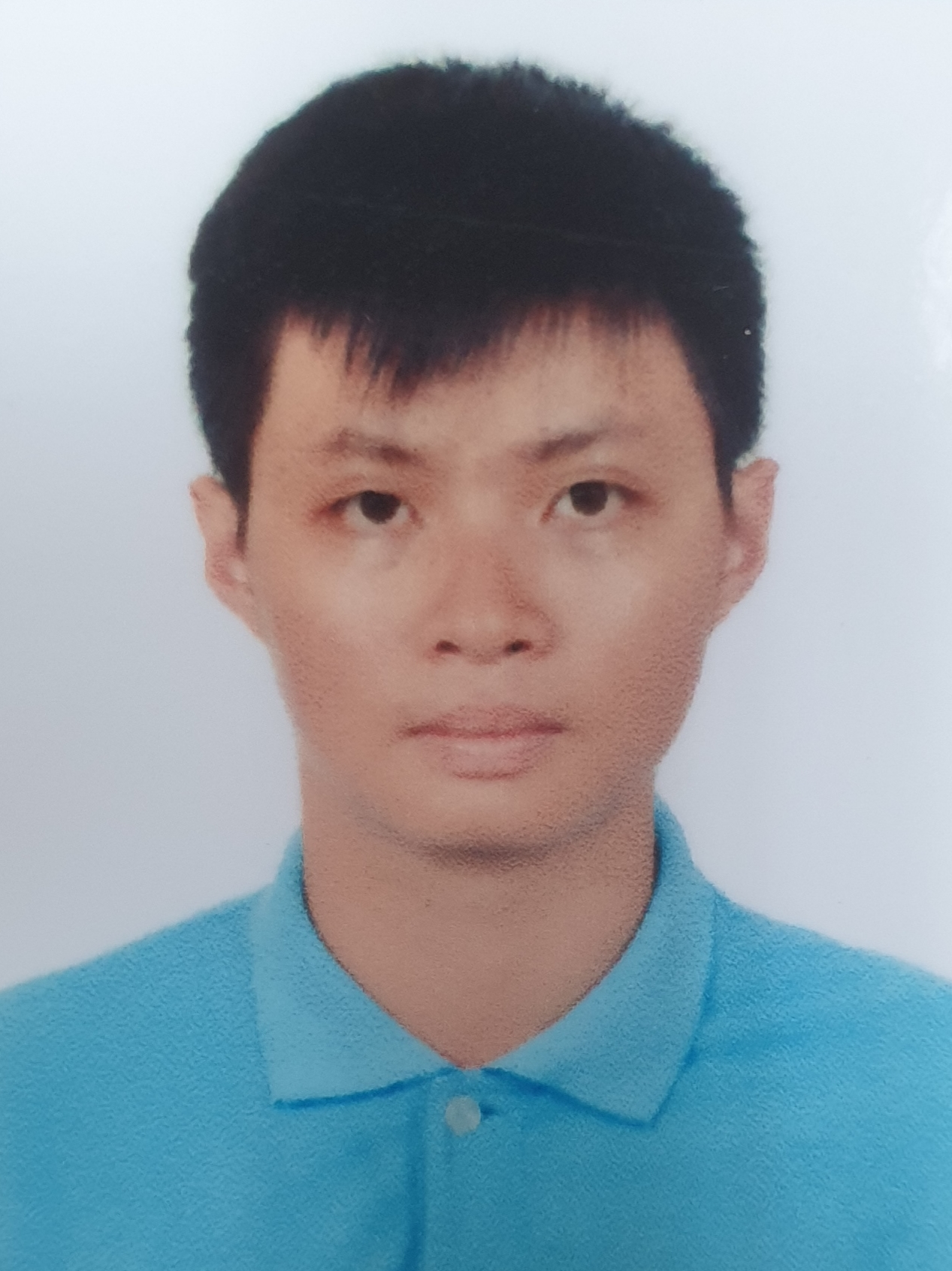}}]{Son T. Luu} received a B.S. degree in 2019 from University of Information Technology, Vietnam National University, Ho Chi Minh City, Vietnam. Currently, he is a research assistant and a master student in the University of Information Technology, Ho Chi Minh City, Vietnam. His research interests include machine reading comprehension, toxic comment detection, sentiment analysis, and knowledge representation. 
\end{IEEEbiography}

\begin{IEEEbiography}[{\includegraphics[width=1in,height=1.25in,clip,keepaspectratio]{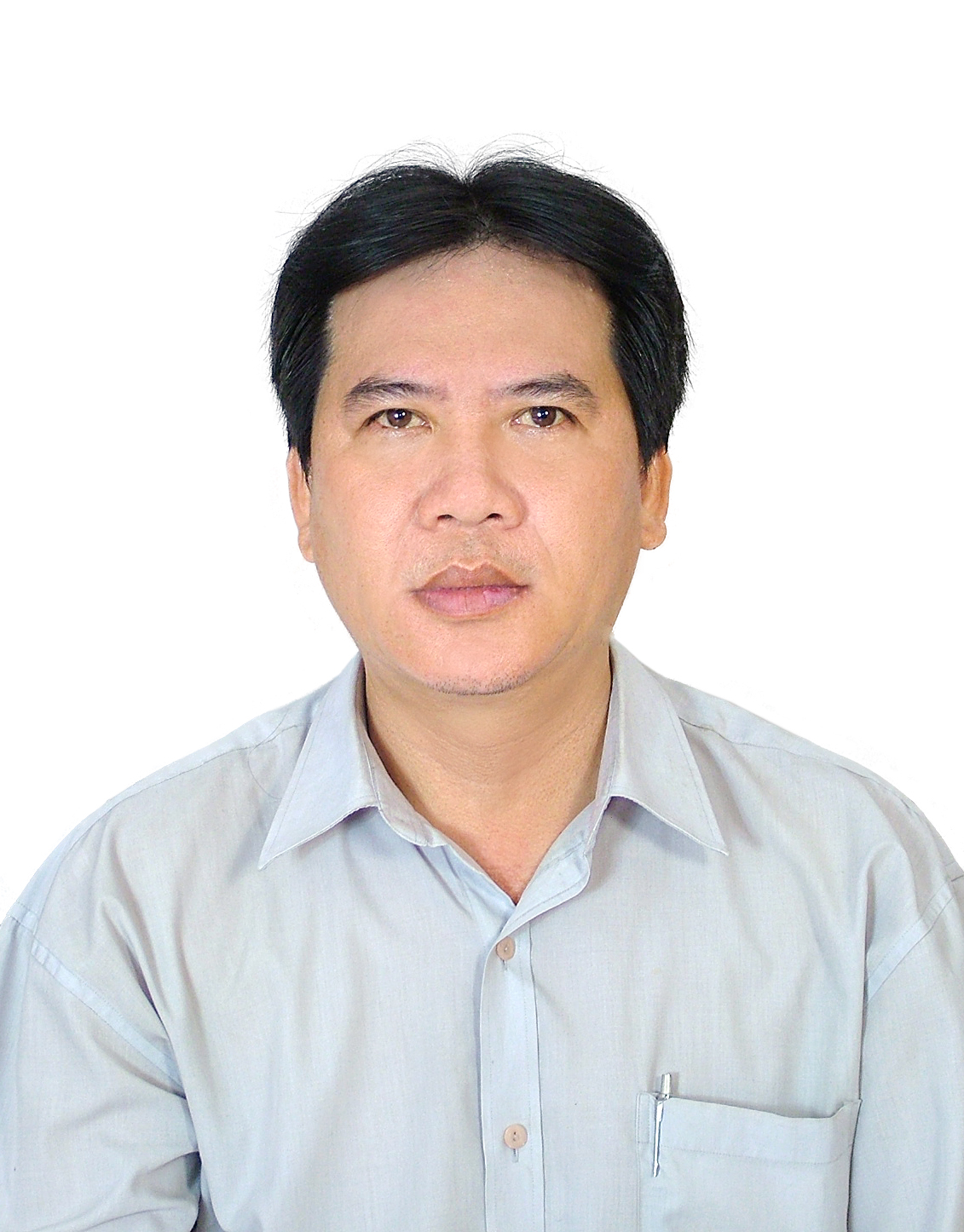}}]{Anh Gia-Tuan Nguyen} is a Dean of Faculty of Information Science and Engineering at University of Information Technology, Vietnam National University, Ho Chi Minh City, Vietnam. He received his B.S., M.S., and PhD. degrees in Information Technology from University of Science, Vietnam National University, Ho Chi Minh City, Vietnam, in 1995, 1998, and 2013, respectively. His research interests include GIS and Intelligent System.
\end{IEEEbiography}

\begin{IEEEbiography}[{\includegraphics[width=1in,height=1.25in,clip,keepaspectratio]{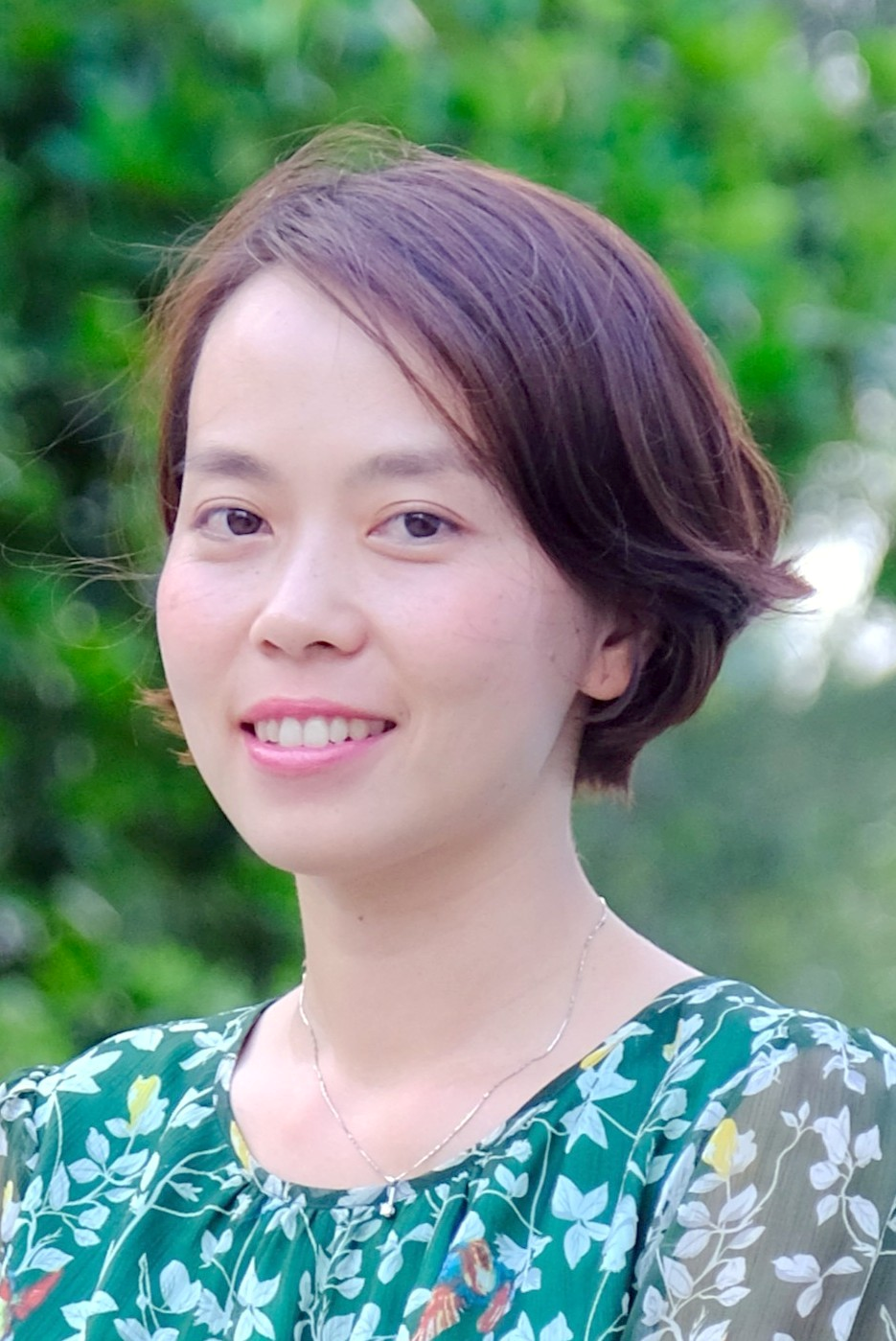}}]{Ngan Luu-Thuy Nguyen} is a scientist at the University of Information Technology, Vietnam National University, Ho Chi Minh City, Vietnam. She received her PhD degree in information science and technology from the University of Tokyo, Japan. She was a postdoctoral researcher at the National Institute of Informatics, Japan from 2012 to 2013. Her research interests include natural language processing and data analysis.
\end{IEEEbiography}

\EOD

\end{document}